\pdfoutput=1
\RequirePackage[svgnames,table]{xcolor}
\documentclass[11pt,letterpaper]{mystyle}

\usepackage[all]{hypcap}
\usepackage[comma,numbers,compress]{natbib}
\usepackage{multirow}
\usepackage{adjustbox}
\usepackage{wrapfig}
\usepackage{placeins}
\usepackage{xspace}
\usepackage{pifont}
\usepackage{array}
\usepackage{longtable}
\usepackage{nicefrac}
\usepackage{xurl}
\usepackage{hyperref}
\usepackage{graphicx}
\usepackage{subcaption}
\usepackage{makecell}
\usepackage{float}
\tcbuselibrary{skins,breakable}

\definecolor{darkblue}{RGB}{196, 86, 18}
\definecolor{TitleBlue}{RGB}{150, 64, 14}
\definecolor{HeaderBlue}{RGB}{176, 76, 16}
\hypersetup{citecolor=darkblue, linkcolor=darkblue, urlcolor=darkblue}

\newcommand{\bench}{\textsc{SciExam for ENSO}\xspace}

\definecolor{BoxBlue}{RGB}{255, 246, 230}
\definecolor{BoxBorder}{RGB}{248, 214, 168}
\definecolor{placeholdergrey}{HTML}{8E8E93}
\definecolor{bestbg}{RGB}{255, 240, 214}
\definecolor{altrow}{RGB}{252, 249, 243}
\definecolor{gaincolor}{RGB}{35, 135, 78}
\definecolor{dropcolor}{RGB}{188, 62, 55}

\definecolor{uparrow}{RGB}{33,102,172}
\definecolor{downarrow}{RGB}{214,96,77}

\newcommand{\cmark}{\textcolor{gaincolor}{\ding{51}}}
\newcommand{\xmark}{\textcolor{dropcolor}{\ding{55}}}
\newcommand{\pmark}{\textcolor{gray}{$\sim$}}

\renewcommand{\titlefont}{\centering\color{TitleBlue}\normalfont\fontsize{18}{20}\selectfont}
\renewcommand{\headingfont}{\color{HeaderBlue}\bfseries\fontsize{13}{14}\selectfont}
\tcbset{titlebox/.style={enhanced, colback=BoxBlue, colframe=BoxBorder, boxrule=0.5mm, arc=2mm, auto outer arc,
    left=5mm, right=5mm, top=5mm, bottom=5mm}}

\renewcommand{\titleboxlogo}{\includegraphics[width=5.5cm]{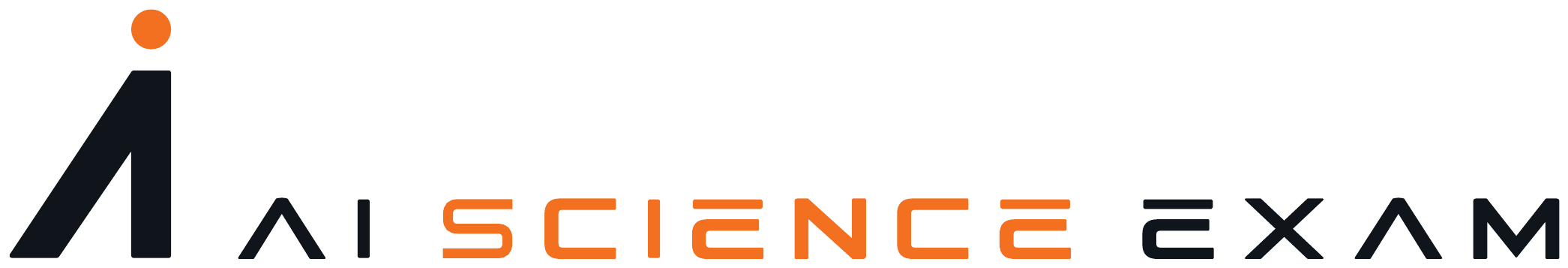}}

\makeatletter
\renewcommand{\maketitle}
{
\vphantom{a}
\vspace{-10mm}
\begin{tcolorbox}[titlebox,
  overlay={\node[anchor=south east, inner sep=0pt] at ([xshift=0cm,yshift=3mm]frame.north east) {\titleboxlogo};}]
\bgroup\setlength{\parindent}{0pt}
  \vskip8pt
  {\centering \titlefont \@title\par}
  \vskip11pt
  {\centering \@author\par}
  \vskip13pt
\egroup
{{\abscontent}}%
\thispagestyle{firststyle}
\end{tcolorbox}
}
\makeatother

\def\Snospace~{Section }

\title{\bench: Can AI Agents Build Climate Models?}
\runningtitle{\bench: Can AI Agents Build Climate Models?}
\author{
  Yinling Zhang$^{1}$, Langchen Liu$^{2}$, Dongbin Xiu$^{1}$, Xueyan Zou$^{3}$, Xu Kuang$^{4}$, Mengdi Wang$^{5}$, Shilong Liu$^{5}$
  \protect\\[2mm]
  {\normalfont\small
  $^1$The Ohio State University \quad $^2$Yale University \quad $^3$University of California, San Diego \quad $^4$Stanford University \quad $^5$Princeton University}
  \protect\\[2mm]
  {\normalfont\small [\href{https://enso.aisexam.com/}{Website}] \quad [\href{https://github.com/ylzhang2447/SciExam-ENSO-code}{Code}]}
}

\date{}

\begin{document}
\begin{abstract}
Language-model agents are increasingly asked to carry out open-ended scientific research, yet their results are usually graded against a known answer, a rubric, or a language-model reviewer, none of which can tell whether a new scientific model is valid. The AI Science Exam for El Ni\~no--Southern Oscillation (\bench{}) is a benchmark in which agents build low-order stochastic models of ENSO, the dominant mode of interannual climate variability, from real observations. Within a six-hour budget, agents process the observations, write their own diagnostics, which are then frozen, and develop a model using only these diagnostics as feedback. Hidden graders then test whether the model reproduces ENSO's statistics, recovers unobserved variables, and forecasts held-out years, and score a published model in the same way. Across twelve agent systems, six produce models that score higher than the published model, mainly through better reconstruction and forecasting. The simplified forms of the stronger models are each compatible with one of the two competing explanations of ENSO's warm–cold asymmetry, an open debate that the task never mentions. Controlled runs of the top system under varied information suggest that its scores do not come from recalling the dated observational record and that the information it receives shapes how it builds its model. \bench{} can thus evaluate agent research where no answer is known, and the results suggest that agents can already build competitive models whose structures bear on questions that scientists still debate.
\end{abstract}
\maketitle
\correspondingauthor{Corresponding author: Shilong Liu, \href{mailto:slongliu86@gmail.com}{slongliu86@gmail.com}}

\begin{figure}[H]
\centering
\includegraphics[width=0.95\linewidth]{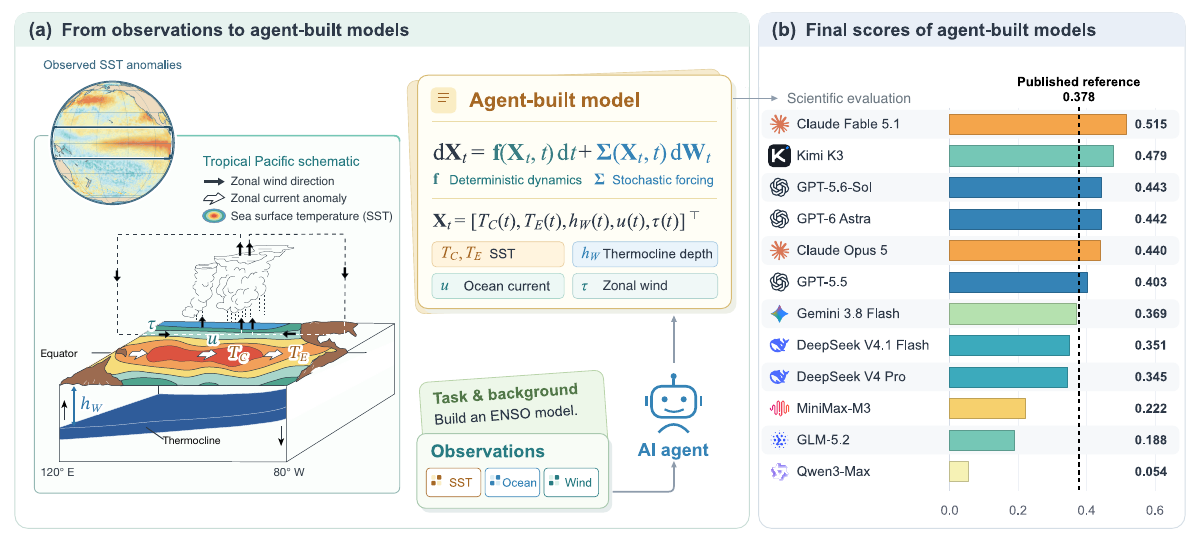}
\caption{ENSO conceptual modeling and benchmark performance. (a) Agents build a stochastic model of five monthly ENSO indices from observations, with deterministic dynamics $\mathbf{f}$ and stochastic forcing $\boldsymbol{\Sigma}$. Schematic adapted from~\citep{fredtheoyster2014enso}, with the five variables marked. (b) Final composite scores of twelve agent systems. The dashed line marks the published reference model (0.378), which six agent-built models exceed.}
\label{figure1_concept}
\end{figure}
\section{Introduction}
\label{sec:intro}
Language-model agents are increasingly evaluated on long-horizon work, from resolving software issues~\citep{jimenez2024swebench} and engineering machine-learning solutions~\citep{chan2025mlebench,wijk2024rebench} to replicating published papers~\citep{starace2025paperbench} and carrying out end-to-end research projects~\citep{lu2024aiscientist,xu2026researchclawbench}, and the tasks they complete autonomously keep growing longer~\citep{kwa2025measuring}. Yet in open-ended scientific research, completing a task is not enough: the result must be scientifically valid and stand up to comparison with existing work.

Existing benchmarks grade agent results, whether answers or written reports, with unit tests or task-specific metrics~\citep{chen2025scienceagentbench}, rubrics derived from a known solution or target findings~\citep{starace2025paperbench,xu2026researchclawbench}, or language-model reviewers~\citep{lu2024aiscientist}. Open-ended research has no single correct answer against which its results can be scored. Dynamical modeling, in which proposed models are simulated and validated against observations, offers a natural testbed for open-ended research. Low-order conceptual models, which reduce a system to a few physically meaningful variables and their interactions (Figure~\ref{figure1_concept}a) rather than simulating it in full detail, make such evaluation especially direct: competing hypotheses about the system can be written down as equations, compared, and tested~\citep{lorenz1963,vialard2025}.

The El Ni\~no--Southern Oscillation (ENSO) is a crucial application of such models. As the dominant mode of interannual climate variability, ENSO affects weather, agriculture, water resources, and marine ecosystems worldwide~\citep{mcphaden2006enso}. Its events recur irregularly, vary in strength and duration, and differ in whether warming peaks in the central or the eastern Pacific~\citep{capotondi2015diversity,timmermann2018}. Although this behavior is hard to capture with a few variables, low-order models have advanced from the recharge oscillator~\citep{jin1997} to recent stochastic models that reproduce much of ENSO's diversity~\citep{chen2022multiscale,zhang2024physics} and predict it competitively~\citep{zhao2024xro}. Yet which processes generate this complexity remains debated~\citep{wang2018,timmermann2018,vialard2025}.

Turning ENSO modeling into a benchmark raises three challenges. First, the evaluation must establish whether a model is scientifically valid without a ground truth. Matching the observed statistics is not enough, because different dynamics can produce similar statistics. Second, the task must give agents enough structure to make progress without prescribing the scientific solution. Given only a broad goal, agents can stall in planning without producing a model, whereas supplying the governing equations would reduce the task to implementation. Third, agents need comprehensive feedback to improve their models over a long run. However, returning grader scores invites agents to exploit weaknesses in the grader~\citep{amodei2016concrete,skalse2022reward}, and partial feedback leads them to neglect unmeasured aspects. Agents must therefore generate their own feedback.

We introduce \bench{}, a benchmark for long-horizon autonomous modeling of ENSO that addresses these challenges with three design choices. First, hidden graders evaluate each model on three complementary aspects: whether its free-running simulations reproduce ENSO's observed statistics, whether it captures variable dependence, and whether it forecasts the held-out years better than persistence. This dependence is tested by reconstructing withheld variables from observed ones. The same graders score a published five-variable model~\citep{zhang2024physics} as a reference, and the first two tests follow the criteria that study used to validate it. Second, the task is divided into observation processing, diagnostic development, and model development, each with an explicit completion condition, within a shared six-hour budget. These stages fix what must be delivered but leave every scientific decision to the agent. Third, agents generate their own feedback by developing diagnostics from the task's scientific guidance. The diagnostics are frozen before model development: agents may execute them to assess each revision but cannot modify them, which prevents their evaluation criteria from drifting toward their results. Grader scores and held-out data remain inaccessible throughout the run.

Six of twelve agent systems produce final models that score higher than the reference model (Figure~\ref{figure1_concept}b), and three more come close. Their advantage lies in reconstruction and forecasting: the reference still reproduces ENSO's statistics best, whereas most agent-built models reconstruct withheld variables and forecast the held-out years more skillfully. Beyond their scores, the models reflect a long-standing scientific debate. ENSO's warm--cold asymmetry has two competing explanations~\citep{martinezvillalobos2019,srinivas2024,vialard2025}, and although the task mentions neither, each model approaching or exceeding the reference can be reduced to a form compatible with one of them. The models thus offer candidate mechanisms rather than mere fits to the data. The shared evaluation also reveals where the two explanations cannot yet be distinguished.

A natural concern is that agents succeed by recalling the observed record rather than modeling it. Controlled changes to the information available to the top system, each run once, show that it still builds a competitive model when the calendar years of the data are hidden. Access to the literature changes how the agent builds its model, toward sparser couplings that follow ENSO theory, while its best score stays similar. More information is not always better: given access to external data, the agent calibrates its model to an operational index instead of the task observations and scores lower.

This work makes three contributions.
\begin{itemize}
  \item \bench{}, a long-horizon benchmark in which agents process real observations, freeze their own diagnostics, and build stochastic ENSO models that hidden graders evaluate against a published reference. Six of twelve agent systems score higher than the reference under the same tests.
  \item A structural analysis showing that the agent-built models approaching or exceeding the reference admit reduced forms in two families, compatible with the two competing explanations of ENSO's warm--cold asymmetry.
  \item Controlled runs of the top system under varied information, which suggest that its scores do not come from recalling the dated observational record and that the information it receives shapes how it builds its model.
\end{itemize}
\section{Related Work}
\label{sec:related}

\begin{table}[t]
\centering
\caption{Comparison with related benchmarks. Long horizon means a budget of an hour or more, physical system means that the task concerns a real physical system, no known answer means grading without reference tests, labels, a paper, an equation, or a fixed solution, hidden scores means that the agent does not see its scores during the run, and scientific evaluation means evaluation beyond fit, such as statistics, coupling, and forecasts. $\sim$ marks partial and a dash unspecified properties.}
\label{tab:related}
\footnotesize
\setlength{\tabcolsep}{6pt}
\begin{tabular}{@{}lccccc}
\toprule
Benchmark & \makecell{Long\\horizon} & \makecell{Physical\\system} & \makecell{No known\\answer} & \makecell{Hidden\\scores} & \makecell{Scientific\\evaluation} \\
\midrule
\multicolumn{6}{@{}l}{\textit{Long-horizon agent tasks}} \\
Agents' Last Exam~\citep{sun2026ale} & \cmark & \xmark & \xmark & \cmark & \xmark \\
MLE-bench~\citep{chan2025mlebench} & \cmark & \xmark & \xmark & \cmark & \xmark \\
RE-Bench~\citep{wijk2024rebench} & \cmark & \xmark & \cmark & \pmark & \xmark \\
ALE-Bench~\citep{imajuku2025alebench} & \cmark & \xmark & \cmark & \xmark & \xmark \\
EdgeBench~\citep{zhu2026edgebench} & \cmark & \pmark & \pmark & \xmark & -- \\
\addlinespace[2pt]
\multicolumn{6}{@{}l}{\textit{Scientific research agents}} \\
PaperBench~\citep{starace2025paperbench} & \cmark & \xmark & \xmark & \cmark & \xmark \\
ScienceAgentBench~\citep{chen2025scienceagentbench} & \xmark & \cmark & \xmark & -- & \xmark \\
ResearchClawBench~\citep{xu2026researchclawbench} & -- & \cmark & \xmark & \cmark & \xmark \\
LLM-SRBench~\citep{shojaee2025llmsrbench} & \xmark & \cmark & \xmark & -- & \xmark \\
\addlinespace[2pt]
\midrule
\rowcolor{BoxBlue}
\textbf{\bench{}} & \cmark & \cmark & \cmark & \cmark & \cmark \\
\bottomrule
\end{tabular}
\end{table}

Long-horizon agent benchmarks test sustained work on software, machine-learning engineering, and optimization tasks~\citep{sun2026ale,chan2025mlebench,wijk2024rebench,imajuku2025alebench,zhu2026edgebench}. Their tasks rarely involve a physical system, and they either grade against a known answer or, when none exists, as in ALE-Bench and most RE-Bench tasks, show the agent its score during the run (Table~\ref{tab:related}). Benchmarks for scientific research agents do involve physical systems, asking agents to replicate papers, complete data-driven analyses, reproduce published findings, or rediscover known equations~\citep{starace2025paperbench,chen2025scienceagentbench,xu2026researchclawbench,shojaee2025llmsrbench}, but each grades the result against a known answer. None tests a result with scientific criteria beyond agreement with a target. In \bench{}, agents build a model of a real physical system, in a task with no known answer, and hidden tests judge its statistics, coupling, and forecasts after the run. Appendix~\ref{app:related} discusses equation discovery and conceptual ENSO models.
\section{Benchmark}
\label{sec:benchmark}

\bench{} mirrors a research setting in which a scientist receives observations, background material, and a modeling objective, and must decide how to build and assess a model. Its design separates the task-specific materials, namely the data, instructions, and graders, from a general execution protocol, so the same structure can be applied to other modeling problems. This study applies the structure to ENSO, where agents must turn raw observations into a calibrated stochastic model.
\subsection{Scientific Modeling Task}
\label{sec:task}

The task is to build a low-order stochastic dynamical model of ENSO that represents variability in both the central and eastern tropical Pacific. The model takes the general form
\begin{equation}\label{eq1}
    \mathrm{d}\mathbf{X}_t
    = \mathbf{f}(\mathbf{X}_t,t)\,\mathrm{d}t
    + \boldsymbol{\Sigma}(\mathbf{X}_t,t)\,\mathrm{d}\mathbf{W}_t,
\end{equation}
where $\mathbf{f}$ describes the deterministic dynamics, $\boldsymbol{\Sigma}$ the stochastic forcing, and $\mathbf{W}_t$ a vector of independent Wiener processes. The state $\mathbf{X}_t = [T_C(t),T_E(t),h_W(t),u(t),\tau(t)]^{\top}$ contains five anomaly variables, namely central and eastern Pacific sea surface temperature $T_C$, $T_E$, western Pacific thermocline depth $h_W$, central Pacific zonal current $u$, and western Pacific zonal wind $\tau$. Together they describe the coupled ocean--atmosphere feedbacks behind ENSO. During El Ni\~no, warm surface anomalies are accompanied by weakened trade winds and a flatter thermocline, and the opposite holds during La Ni\~na.

\subsection{Agent Workflow}
\label{sec:workflow}
\begin{figure}[h]
\centering
\includegraphics[width=1\linewidth]{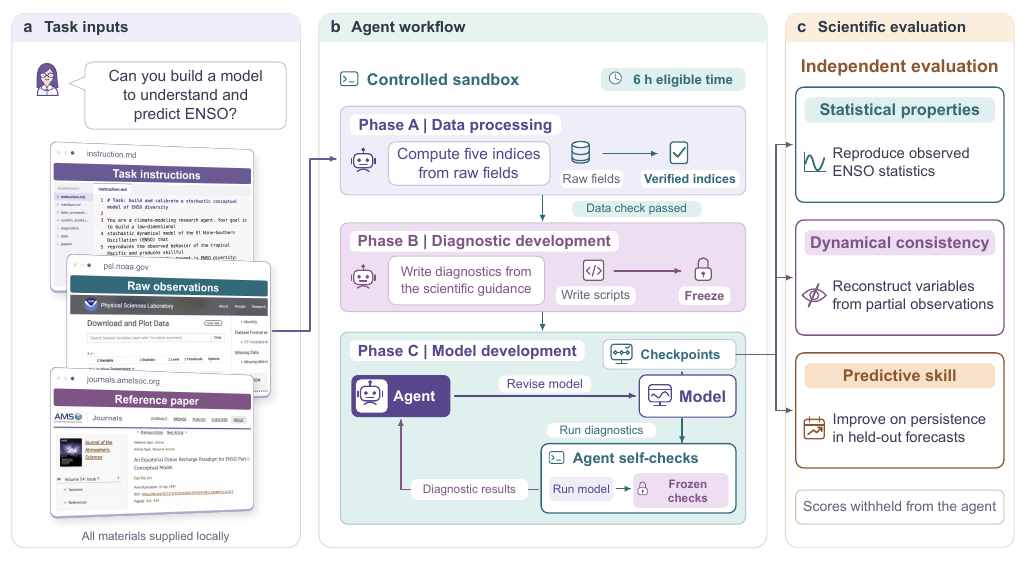}
\caption{Overview of \bench{}. Given background material, raw observations, and a reference paper, the agent works through three phases within a six-hour budget: (A) processing the observations into indices that must pass a data check, (B) writing its own diagnostics from the task's scientific guidance, which are then frozen, and (C) developing the model while rerunning these diagnostics. Hidden graders evaluate the final model on statistical properties, dynamical consistency, and predictive skill without returning scores to the agent. Dialogue is illustrative.}
\label{fig1_overview}
\end{figure}

Agents work through three phases that share a six-hour budget, as shown in Figure~\ref{fig1_overview}. The phases follow the order in which a scientist would approach the problem, first preparing the data, then deciding how to judge a model, and finally building one. This structure fixes what must be delivered while leaving every scientific decision to the agent.

In Phase A, agents process the raw observations into the five monthly indices, and modeling begins only after their indices match a correctly processed version. This gate keeps processing errors out of the model and ensures that all agents model the same data, so their scores are directly comparable.

In Phase B, agents write their own diagnostics, choosing which quantities to check and how, from instructions that state what the final model will be judged on but not how it is measured. The phase ends when the agent declares the diagnostics complete or after 90 minutes. Checks of their own let agents judge each revision without seeing the graders' scores, which would invite tuning to the scores rather than to the science.

In Phase C, agents build and calibrate the model, rerunning their diagnostics to judge each revision. The diagnostics are frozen at the start of this phase, so they are the only feedback agents receive and their criteria cannot drift toward what the model already does.

\subsection{Scientific Evaluation}
\label{sec:evaluation}
Each model receives a score from $0$ to $1$ on three dimensions (Table~\ref{tab:evaluation}), and its composite score is their weighted sum. All three are computed by running the same submitted model, and the published reference model~\citep{zhang2024physics} is scored in the same way. Each test uses the period suited to it: a long record (1950--2019) for statistics, the reliably observed years 1980--2014 for reconstruction, and years the agents never saw (2015--2024) for prediction. The agents receive only the 1980--2014 data, whereas the reference was developed with observations covering all three periods and was not tuned to this composite score. The metrics and their formulas are given in Appendix~\ref{app:evaluation}.

\begin{table}
\centering
\caption{The three evaluation dimensions. Each is scored from $0$ to $1$ by running the submitted model, and the composite score is their weighted sum. The last column gives the panels of Figure~\ref{fig3_final_diagnostics} that show each measure for the top-scoring model. Persistence, the forecast baseline, keeps the initial anomaly unchanged.}
\label{tab:evaluation}
\small
\renewcommand{\arraystretch}{1.15}
\setlength{\tabcolsep}{4.5pt}
\begin{tabularx}{\linewidth}{@{}>{\raggedright\arraybackslash}p{0.13\linewidth}>{\raggedright\arraybackslash}p{0.31\linewidth}>{\raggedright\arraybackslash}X>{\centering\arraybackslash}p{0.14\linewidth}@{}}
\toprule
Dimension & What it tests & Measure & Panels \\
\midrule
\multirow{5}{=}{Statistical properties} & \multirow{5}{=}{Whether a long free run of the model behaves like the observed ENSO}
  & Probability density functions (PDFs) & a, e \\
& & Seasonal variance & b, f \\
& & Autocorrelation functions (ACFs) & c, g \\
& & Frequency of different event types & d \\
& & Event strength against location & h \\
\midrule
\multirow{2}{=}{Dynamical consistency} & \multirow{2}{=}{Whether the variables are coupled as observed}
  & $h_W$, $u$, $\tau$ reconstructed from $T_C$, $T_E$ & i--k \\
& & $T_C$, $T_E$ reconstructed from $h_W$, $u$, $\tau$ & l, m \\
\midrule
\multirow{2}{=}{Predictive skill} & \multirow{2}{=}{Whether the model forecasts years it has never seen}
  & Forecasts at lead times of 1--12 months compared with persistence & n, o \\
\bottomrule
\end{tabularx}
\end{table}

\subsection{Experimental Setup}
\label{sec:setup}
We evaluate twelve agent systems, each pairing a language model with an agent scaffold: Claude Code, Codex, and Antigravity for Claude, GPT, and Gemini models, respectively, and Claude Code connected to OpenRouter for the others (Appendix~\ref{app:harness}). Each system runs in its own offline container, and the harness saves a checkpoint every five minutes of eligible time, which excludes waits caused by the model provider. To encourage exploration, agents are told that their best saved submission counts, but systems are compared on their final checkpoints. Best checkpoints would score higher, with eight rather than six systems exceeding the reference (Appendix~\ref{app:harness}). Appendix~\ref{app:task} summarizes the task materials and Appendix~\ref{app:infra} reports the cost of the runs.

\section{Results}
\label{sec:results}

\begin{figure}[h]
    \centering
    \includegraphics[width=1\linewidth]{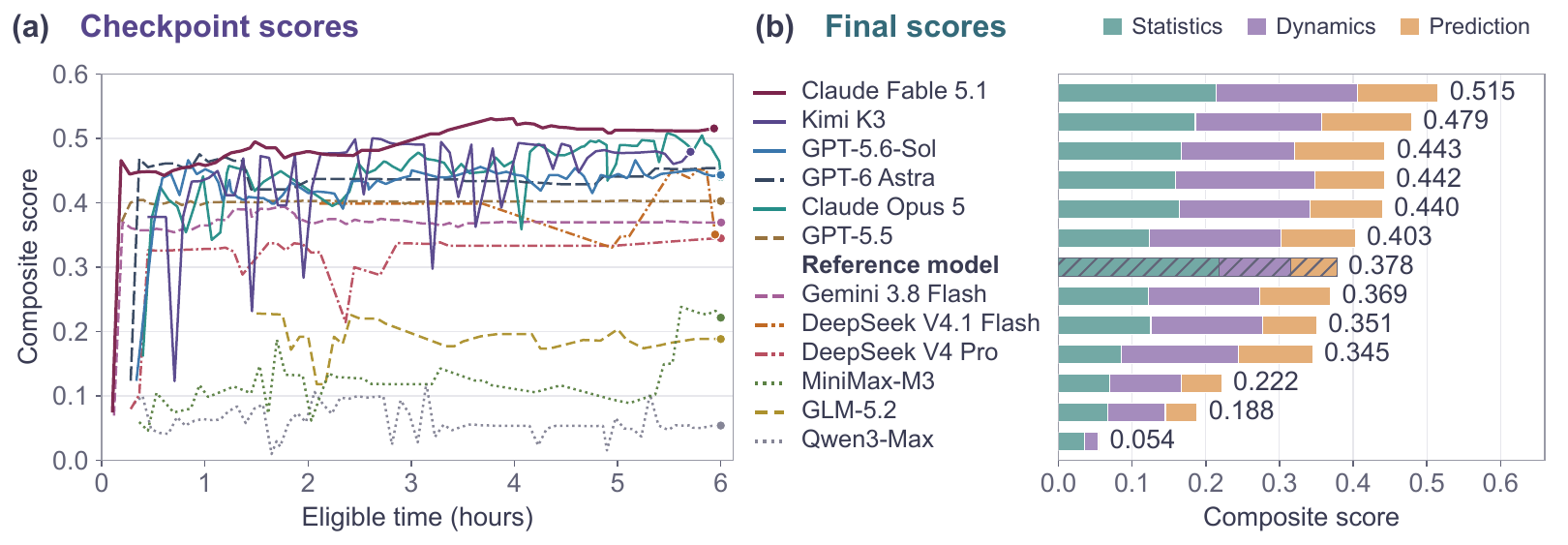}
    \caption{Model development and final scores. Panel (a): Composite scores of the checkpoints of each system against eligible time from the start of the run, including data processing and diagnostic development. Filled circles mark final checkpoints, and colors and line styles follow the system names in panel (b). Panel (b): Final scores of the twelve systems and the reference model, ranked by composite score. Stacked segments show the weighted contributions of statistical properties, dynamical consistency, and predictive skill, with weights of $0.3$, $0.3$, and $0.4$, and numbers at the ends give the composite scores.}
    \label{fig2_checkpoint_and_final_scores}
\end{figure}

Figure~\ref{fig2_checkpoint_and_final_scores} shows how the model of each system develops during the run and how its final model scores. 
In panel (a), progress within a run is not monotonic, partly because agents keep exploring alternative model structures rather than only refining the best one found so far. Claude Opus 5 and DeepSeek V4.1 Flash, for example, reach their highest scores late in the run but submit final models that score below those checkpoints.
Six of the twelve final models exceed the composite score of the published reference model ($0.378$), led by Claude Fable 5.1 ($0.515$), Kimi K3 ($0.479$), and GPT-5.6-Sol ($0.443$), and three more score close to it, between $0.345$ and $0.369$. Differences of about $0.01$, as among GPT-5.6-Sol, GPT-6 Astra, and Claude Opus 5, should not be read as a ranking. The reference still reproduces the observed statistics best ($0.727$), whereas most agent-built models score higher on dynamical consistency and predictive skill, on which the reference scores $0.323$ and $0.157$. They do so even though the reference was developed with observations from all three test periods (Section~\ref{sec:evaluation}). Agents thus build, from only 35 years of observations, models that rival a published one.

\begin{figure}[h]
    \centering
    \includegraphics[width=0.95\linewidth]{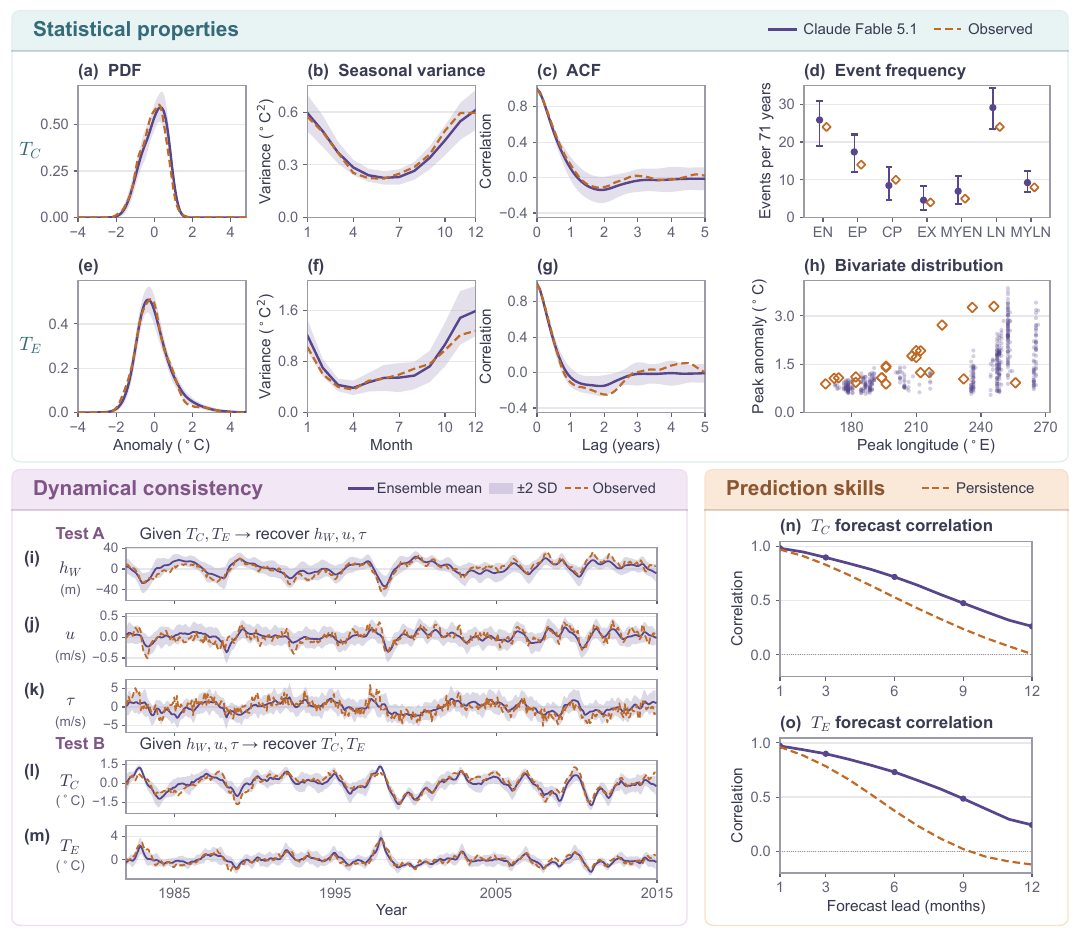}
    \caption{Scientific diagnostics of the final ENSO model produced by Claude Fable 5.1. Panels (a--c, e--g): Simulated and observed probability density functions (PDFs), seasonal variance, and autocorrelation functions (ACFs) for $T_C$ and $T_E$. Panel (d): Event counts per seventy-one years, with model means and 2.5th--97.5th percentile ranges compared against observed counts. EN and LN denote El Ni\~{n}o and La Ni\~{n}a. EP and CP denote eastern- and central-Pacific El Ni\~{n}o, and EX denotes extreme El Ni\~{n}o. MYEN and MYLN denote multi-year events. Panel (h): Peak temperature anomalies versus peak longitudes for simulated and observed events. Panels (i--m): State reconstruction from partial measurements. Test A uses $T_C$ and $T_E$ to reconstruct $h_W$, $u$, and $\tau$. Test B swaps the observed and reconstructed variables. Panels (n, o): Correlations between ensemble-mean forecasts and observations at lead times of one to twelve months over the held-out 2015--2024 period, compared with persistence.}
\label{fig3_final_diagnostics}
\end{figure}

The final model from Claude Fable~5.1 nearly matches the reference in statistics while scoring clearly higher on the other two dimensions (Figure~\ref{fig3_final_diagnostics}). Its simulations reproduce the contrasting distributions, seasonal variance, and autocorrelation of $T_C$ and $T_E$ (panels a--c, e--g), stronger eastern than central Pacific events (h), and observed counts of all seven event types within the simulated $95\%$ ranges (d). It recovers the withheld variables in both reconstruction directions (i--m) and forecasts both indices better than persistence at all twelve lead times (n, o), so a single agent-built model can perform well on all three dimensions at once.

\section{Analysis}
\label{sec:analysis}

Section~\ref{sec:results} showed that several agents build models that exceed a published reference under the same evaluation. This section asks which model structures carry this performance across agents, since they may offer insight into ENSO itself, and how access to external information changes the way a single agent models ENSO.

\subsection{Agent-Built Models Reenact the ENSO Mechanism Debate}
\label{sec:analysis_mechanism}

Which processes generate ENSO's complexity remains debated. Conceptual theories differ in which ocean and atmosphere processes drive the oscillation and in whether ENSO is self-sustained or a damped mode excited by stochastic forcing~\citep{wang2018,timmermann2018,vialard2025}. A prominent open question concerns its asymmetry, as warm events in the eastern Pacific tend to be stronger than cold events and give sea surface temperature a positively skewed distribution. One explanation attributes this asymmetry to deterministic nonlinearity in the coupled dynamics, such as a nonlinear wind response to sea surface temperature~\citep{srinivas2024}. Another attributes it to multiplicative noise, stochastic forcing whose amplitude depends on the state, with which even a linear model reproduces the observed asymmetry~\citep{martinezvillalobos2019}. Both are physically plausible, and their relative contributions remain unresolved~\citep{bianucci2018,vialard2025}.

\FloatBarrier

The agent-built models offer an independent view of this question, since none of the agents was told of either explanation. Each final model, however, mixes the terms that drive its behavior with components added during development and never removed. To find which terms matter, groups of terms were removed from each final model after all runs had ended, the remaining coefficients were refitted, and the reduced model was rescored with the same graders. A reduction was kept only if it retained nearly all of the original scores and the asymmetry of the model's temperature distribution (Appendix~\ref{app:simplification}). The analysis covers the nine models that exceed or approach the reference and asks which explanation of the asymmetry each reduced model relies on.

\begin{figure}[h]
    \centering
    \includegraphics[width=1\linewidth]{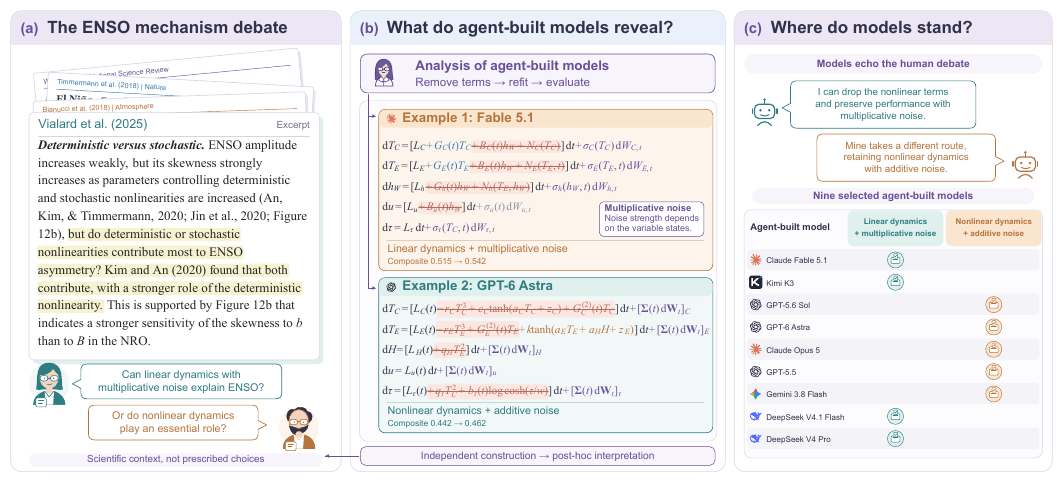}
    \caption{Structural simplification of agent-built ENSO models. (a) Nonlinear dynamics and stochastic forcing as two established explanations of ENSO asymmetry, in an excerpt from \citet{vialard2025}. (b) Two ablation examples. $L_i$ collects constant and linear coupling terms, $G_i(t)$ is the seasonal growth or damping rate of the variable itself ($G_i^{(2)}$ its semiannual part), $B_i(t)$ is the seasonal coupling to $h_W$, $N_i$ denotes nonlinear terms, and $H$ is an internal heat-content variable of GPT-6 Astra mapped to $h_W$. Other letters are parameters, seasonal when written with $t$. $dW$ denotes increments of a Wiener process, scaled by the state-dependent $\sigma_i$ in Claude Fable~5.1 (multiplicative noise) and by the time-dependent matrix $\Sigma(t)$ in GPT-6 Astra (additive noise). (c) Classification of the nine models that exceed or approach the reference by their higher-scoring viable reduction, with no family assigned in advance, recovering the two model families debated for ENSO.}
    \label{fig5_ablation_mechanism_story}
\end{figure}

The simplified models divide in how they generate the asymmetry of ENSO (Figure~\ref{fig5_ablation_mechanism_story}b). Claude Fable~5.1 loses all of its nonlinear drift terms and reduces to linear drift with multiplicative noise, while its score rises from $0.515$ to $0.542$. GPT-6 Astra instead retains a nonlinear temperature feedback driven by additive noise, with its score rising from $0.442$ to $0.462$. Across the nine models, the simplified forms fall into these two families, four with linear drift and multiplicative noise and five with nonlinear drift and additive noise, matching the two explanations of ENSO asymmetry (Figure~\ref{fig5_ablation_mechanism_story}c). The agent-built models thus admit reduced forms compatible with both established explanations. Simplification costs the models little. None of the nine loses more than $0.014$, and statistical scores often rise at a small cost in dynamical consistency (Table~\ref{tab:simplification}), so many terms added during development contributed little. These gains should be read with caution, because the same graders selected the reductions. GPT-5.5 can be simplified into either family, and the two reductions score within $0.3$ percentage points of each other ($0.398$ and $0.395$), so the evaluation cannot tell which explanation its dynamics require. The same ambiguity keeps the question open in the literature.

\begin{table}[!htb]
\centering
\newcommand{\dup}[1]{\,{\scriptsize\textcolor{gaincolor}{($\uparrow$#1)}}}
\newcommand{\ddn}[1]{\,{\scriptsize\textcolor{dropcolor}{($\downarrow$#1)}}}
\newcommand{\dzero}{\,{\scriptsize\textcolor{gray}{(0.000)}}}
\caption{Structure and scores of the nine analyzed models before and after simplification. Drift is nonlinear (NL) or linear (L), and noise is multiplicative (Mult.) or additive (Add.). Parentheses give the change from the original model (green for an increase, red for a decrease), and bold marks the higher value for each model. GPT-5.5 can be simplified into either family.}
\label{tab:simplification}
\small
\setlength{\tabcolsep}{3.5pt}
\begin{adjustbox}{max width=\linewidth}
\begin{tabular}{@{}llccllll@{}}
\toprule
\multirow{2}{*}{Model} & \multirow{2}{*}{Version} & \multicolumn{2}{c}{Structure} & \multicolumn{3}{c}{Component scores} & \multirow{2}{*}{Composite} \\
\cmidrule(lr){3-4}\cmidrule(lr){5-7}
 & & Drift & Noise & Stat. & Dyn. & Pred. & \\
\midrule
\multirow{2}{*}{Claude Fable~5.1} & Original & NL & Mult. & 0.715 & \textbf{0.639} & 0.274 & 0.515 \\
 & Simplified & L & Mult. & \textbf{0.810}\dup{0.095} & 0.600\ddn{0.039} & \textbf{0.297}\dup{0.023} & \textbf{0.542}\dup{0.027} \\
\addlinespace[2pt]
\multirow{2}{*}{Kimi~K3} & Original & NL & Mult. & \textbf{0.620} & 0.570 & \textbf{0.305} & \textbf{0.479} \\
 & Simplified & L & Mult. & 0.575\ddn{0.045} & \textbf{0.573}\dup{0.003} & 0.302\ddn{0.003} & 0.465\ddn{0.014} \\
\addlinespace[2pt]
\multirow{2}{*}{GPT-5.6-Sol} & Original & NL & Add. & 0.554 & 0.515 & \textbf{0.306} & 0.443 \\
 & Simplified & NL & Add. & \textbf{0.613}\dup{0.059} & \textbf{0.529}\dup{0.014} & 0.304\ddn{0.002} & \textbf{0.464}\dup{0.021} \\
\addlinespace[2pt]
\multirow{2}{*}{GPT-6 Astra} & Original & NL & Add. & 0.530 & \textbf{0.629} & 0.237 & 0.442 \\
 & Simplified & NL & Add. & \textbf{0.598}\dup{0.068} & 0.611\ddn{0.018} & \textbf{0.248}\dup{0.011} & \textbf{0.462}\dup{0.020} \\
\addlinespace[2pt]
\multirow{2}{*}{Claude Opus~5} & Original & NL & Mult. & 0.548 & \textbf{0.589} & 0.247 & 0.440 \\
 & Simplified & NL & Add. & \textbf{0.629}\dup{0.081} & 0.567\ddn{0.022} & \textbf{0.286}\dup{0.039} & \textbf{0.473}\dup{0.033} \\
\addlinespace[2pt]
\multirow{3}{*}{GPT-5.5} & Original & NL & Mult. & \textbf{0.412} & 0.595 & 0.252 & \textbf{0.403} \\
 & Simplified & NL & Add. & 0.390\ddn{0.022} & \textbf{0.600}\dup{0.005} & 0.251\ddn{0.001} & 0.398\ddn{0.005} \\
 & Simplified & L & Mult. & 0.387\ddn{0.025} & 0.590\ddn{0.005} & \textbf{0.255}\dup{0.003} & 0.395\ddn{0.008} \\
\addlinespace[2pt]
\multirow{2}{*}{Gemini~3.8 Flash} & Original & NL & Mult. & \textbf{0.406} & \textbf{0.503} & 0.241 & \textbf{0.369} \\
 & Simplified & NL & Add. & 0.389\ddn{0.017} & 0.499\ddn{0.004} & 0.241\dzero & 0.362\ddn{0.007} \\
\addlinespace[2pt]
\multirow{2}{*}{\makecell[l]{DeepSeek\\V4.1 Flash}} & Original & NL & Mult. & 0.418 & 0.508 & 0.183 & 0.351 \\
 & Simplified & L & Mult. & \textbf{0.419}\dup{0.001} & \textbf{0.559}\dup{0.051} & \textbf{0.244}\dup{0.061} & \textbf{0.391}\dup{0.040} \\
\addlinespace[2pt]
\multirow{2}{*}{\makecell[l]{DeepSeek\\V4 Pro}} & Original & NL & Mult. & \textbf{0.285} & 0.530 & \textbf{0.251} & \textbf{0.345} \\
 & Simplified & L & Mult. & 0.269\ddn{0.016} & \textbf{0.532}\dup{0.002} & 0.250\ddn{0.001} & 0.340\ddn{0.005} \\
\bottomrule
\end{tabular}
\end{adjustbox}
\end{table}

\subsection{A Case Study of the Top System: Memorized and External Knowledge}
\label{sec:analysis_conditions}

Additional six-hour runs of Claude Fable~5.1, the highest-scoring system, tested where its performance came from, with full scores in Appendix~\ref{app:fable}. Each run changed one condition of the formal offline run. The first stayed offline but replaced every calendar year in the data and task materials with a relative label. The other two allowed web access, first to both the literature and data sources and then to the literature only, with the publication describing the reference model blocked in both. Each condition was run once, so these comparisons describe one agent under controlled changes rather than effects estimated across repeated runs.

\paragraph{Robustness to data contamination.} Language models may succeed by recalling test data from their training corpus rather than by solving the task, and ENSO observations and event histories are widely published. Such recall helps little here, because a model with a few dozen coefficients cannot store an observed record, and knowing when past events occurred does not yield coefficients that reproduce ENSO's statistics, dynamics, and forecasts. When each run is scored under the calendar convention its agent stated, the anonymized run reaches a best score of $0.482$ and a final score of $0.463$, compared with $0.531$ and $0.515$ in the formal run, and it still reproduces the main statistical features of ENSO. A competitive model can therefore be built without knowing which years were observed.

\paragraph{Effect of web access.} More information did not always help. With data sources available, the agent adopted the event list of an operational ENSO index as its calibration target in place of the task data and reached the lowest best score of the four conditions ($0.475$), so data sources were excluded from the next run. With the literature only, the agent reached a best score of $0.528$, close to the offline $0.531$, but built its model differently. The offline and literature-only agents both began with a process-based recharge oscillator. The offline agent abandoned it after its first hand-set calibration proved unstable and fitted a dense linear matrix with all twenty-five couplings. The web agent kept the process structure and included only couplings with a physical pathway, whose signs match ENSO theory, including the recharge cycle and the Bjerknes feedback. This sparser model scores higher on statistics ($0.798$ versus $0.741$) and lower on dynamical consistency ($0.556$ versus $0.632$), the same trade-off that simplification shows in Section~\ref{sec:analysis_mechanism}. Its simplified core is closest to the linear family. All explicit nonlinear corrections can be removed without loss except a cubic damping added for stability, whereas replacing multiplicative with additive noise costs $12\%$ of the composite score, about twice the $6\%$ lost by removing the damping. Both web runs ended well below their best checkpoints, with final scores of $0.386$ and $0.406$, a much larger decline than offline.

\section{Conclusion}
\label{sec:conclusion}

\bench{} evaluates agents that develop stochastic ENSO models from real observations, using hidden, independent scientific tests that require no known answer. Under these tests, six of twelve agent systems exceed a published reference model, and the simplified forms of the stronger models are each compatible with one of the two debated explanations of ENSO asymmetry. Agents can therefore already contribute candidate models to open scientific questions.

\FloatBarrier
\bibliographystyle{plainnat}
\bibliography{refs_preprint}

\appendix
\section{Task Materials}
\label{app:task}

This appendix summarizes what each agent receives. The task files and every prompt that the harness sends are included verbatim in the released code and data.

\paragraph{Inputs.}
Agents receive raw gridded observations from January 1980 to December 2014, namely sea surface temperature from ERSSTv5~\citep{huang2017ersst}, ocean potential temperature and zonal currents from GODAS~\citep{behringer2004godas}, and 850\,hPa zonal wind from the NCEP--NCAR reanalysis~\citep{kalnay1996ncep}. A prescribed procedure, summarized below, specifies how to build the five monthly anomaly indices from these fields, and observations from 2015 onward are withheld. As background, agents receive the recharge-oscillator paper of \citet{jin1997}, which links eastern Pacific sea surface temperature and western Pacific thermocline depth in a two-variable model with linear dynamics and simple nonlinear and stochastic extensions. They also receive characteristic scales for nondimensionalizing the indices.

\paragraph{Submission.}
Within the general form of Eq.~(\ref{eq1}), agents decide the governing equations themselves, including how the variables interact and how nonlinearity, seasonal dependence, and stochastic forcing are represented. Each submission is an executable implementation of Eq.~(\ref{eq1}) with its calibrated parameters, accompanied by a model card that describes the equations and the mechanisms they represent. The graders run this model to generate the simulations, reconstructions, and forecasts on which it is scored.

\paragraph{Task instruction.}
The instruction file asks the agent to build a low-dimensional stochastic dynamical model of ENSO that represents both central and eastern Pacific variability, reproduces the observed behavior of the tropical Pacific, and produces skillful forecasts. It describes a data-processing stage whose output must pass a check before modeling is unlocked, and a modeling stage in which the agent chooses the equations, nondimensionalizes the indices with the provided scales, calibrates the parameters, and compares the model with statistics computed from its own processed data. It names the scoring categories, namely the statistics of the simulated climate, the physical mechanism, forecast skill relative to persistence on initialization months the agent never sees, and a validity gate under which an unstable or non-finite model scores zero, without giving thresholds, weights, or held-out data. It also states the rules. The agent must build the model itself, must not use publications that already contain a complete calibrated model of this system, and must stay inside its working directory without accessing any grader, reference model, or observation file outside it.

\paragraph{Data processing.}
The processing file prescribes a fixed recipe, and all regional averages are weighted by the cosine of latitude over boxes between $5^{\circ}$S and $5^{\circ}$N. $T_C$ and $T_E$ are sea surface temperature averages over the Ni\~no~4 box, from $160^{\circ}$E to $150^{\circ}$W, and the Ni\~no~3 box, from $150^{\circ}$W to $90^{\circ}$W. $h_W$ is the depth of the $20^{\circ}$C isotherm averaged from $120^{\circ}$E to $180^{\circ}$, where the depth in each water column is found by linear interpolation between the two GODAS levels that bracket the first crossing from the surface. $u$ is the zonal current averaged over the levels from the surface to $55$\,m and over the Ni\~no~4 box. $\tau$ is the $850$\,hPa zonal wind averaged from $120^{\circ}$E to $180^{\circ}$, first over the box for each day and then over each month, with westerly winds positive. Anomalies are formed by removing the monthly climatology of 1981 to 2010 from every index, without detrending. The file also lists common pitfalls, such as fill values, land masks, longitudes across the dateline, and GODAS temperatures given in Kelvin. The data check compares each index element-wise with a reference processed in the same way, within a tight tolerance. When the check fails, the feedback names the indices that differ and the month of the largest discrepancy.

\paragraph{Model interface.}
The interface file fixes the form of the submission. The submitted model exposes a single time-stepping function, which advances an ensemble of five-variable states, given in physical units, by one stochastic step of a given length in months at a given model time, drawing all random numbers from a generator supplied by the graders. The graders build every experiment on this function, namely free simulations from the zero state, ensemble forecasts initialized from the observed indices, and an ensemble Kalman smoother that observes some variables and reconstructs the others, which the file describes as a test of the dependence among variables. A submission whose model fails to import, raises an error, or returns non-finite values fails the validity gate. The calibrated parameters are stored in a separate parameter file, and the model card describes the state variables, equations, and mechanisms.

\paragraph{Scales and excluded references.}
The scales file gives $7.5\,^{\circ}$C for the temperatures, $150$\,m for $h_W$, $1.5$\,m\,s$^{-1}$ for $u$, $5$\,m\,s$^{-1}$ for $\tau$, and two months for time, and states that these are unit conventions that do not prescribe any model structure. The list of excluded references covers publications that already contain a complete calibrated model of the five-variable system, including the published reference model~\citep{zhang2024physics} and the review of \citet{vialard2025}, whose tables give the full nonlinear stochastic recharge oscillator. A note accompanying the paper of \citet{jin1997} states that it contains only the linear deterministic core, and that the stochastic forcing, nonlinearity, seasonal dependence, and ENSO diversity are left to the agent.

\paragraph{Phase prompts.}
The harness sends one prompt at the start of each phase. The Phase A prompt asks the agent to process the raw data into the five indices by following the processing file, and a failed check returns the prompt with the feedback, for up to four attempts. The Phase B prompt tells the agent that nothing in its workspace will score its model and asks it to write diagnostic scripts that run the submitted model and report how far it is from the observations. It asks each script to report the quantity, the observed value, and the gap rather than a pass or fail verdict, and to state the discrepancy it finds. It warns that the diagnostics become read-only once the agent stops, and it describes the three evaluation dimensions in one sentence each, as whether a long free run looks like the observed climate, whether the model behaves as a coupled dynamical system, and whether it forecasts better than persistence from an observed initial state, adding that these are descriptions rather than metrics. If a script fails to run, a repair prompt asks the agent to fix or delete it before the diagnostics are frozen. The Phase C prompt asks for the three submission files, states that the diagnostics are now read-only and that no new evaluation code should be written, warns that agreement between the checks and the observations does not mean the model is finished, and repeats the three descriptions. To encourage exploration, it also explains that the submission is copied automatically and that the best copy counts, so that a failed experiment cannot lower the grade. It further notes that running simulations consumes the same time budget and that the agent should keep working for the whole budget.

Later turns of Phase C use a resume prompt that restates these points. After 20, 40, and 60 minutes without a new copy of the submission, escalating prompts ask the agent to run its diagnostics, find its weakest aspect, and change the model, and to reconsider its structure if parameter tuning is exhausted. In Claude Code, a stop hook blocks a voluntary exit with a short reminder to check the model on the three dimensions, until the submission has stayed unchanged over three stop attempts, and the Gemini scaffold receives the same reminder through its continuation loop. Every prompt ends with a note asking the agent to keep a concise research log in its workspace, because its working context may be compacted.
\section{Harness Details and Reproducibility}
\label{app:harness}

\subsection{Evaluation Protocol}
\label{sec:protocol}

\paragraph{Agent systems.}
Each evaluated system pairs a language model with an agent scaffold, the software through which the model reads files, runs code, and manages its own work. Because the scaffold shapes how a model works over a long run, the benchmark compares complete systems rather than models alone. Claude, GPT, and Gemini models run in the scaffolds their developers provide for agentic work, namely Claude Code, Codex, and Antigravity, each through its subscription service, while the remaining models run in Claude Code connected to OpenRouter through a LiteLLM gateway. Each system runs at the settings of its service: medium reasoning effort for subscription systems and the provider default for the rest, so differences between systems reflect the model together with its scaffold and settings rather than the model alone.

\paragraph{Environment.}
Each run takes place in its own container, built from the same image and mounting only the working directory of that run. The graders, the reference model, and the held-out observations stay outside the container, so everything an agent knows about the task comes from the materials it is given and its own work. The isolation measures are described below.

\paragraph{Time budget.}
Each run has a six-hour budget of eligible time shared across the three phases. This is long enough for every system to finish while keeping full runs of twelve systems practical. Following earlier long-horizon benchmarks~\citep{wijk2024rebench,starace2025paperbench}, eligible time excludes waits caused by the model provider, namely exhausted usage quotas and transient service failures, so that every system is compared over the same amount of working time.

\paragraph{Runs and checkpoints.}
Each system is run once. The object of evaluation is the model that each run produces, which is scored and inspected as a scientific result in its own right, so the comparisons describe these models rather than the expected performance of each system. During model development, the harness copies the submission every five minutes of eligible time, at the end of every agent turn, and once more when the run ends, and each copy that differs from all earlier copies is kept as a checkpoint. All checkpoints are scored after the run, which shows how the model develops over time. The prompts tell the agent that its best copy counts, which is intended to let it explore without protecting a working version. The main comparison nonetheless uses the final checkpoint, so it differs from the objective stated to the agent. The best checkpoint can only be identified by scoring all checkpoints with the hidden graders, which selects on the test metric itself and is not something the agent can do. The final checkpoint is the model the agent had at the end of its budget, and because the best checkpoint always scores at least as high, this choice can only understate each run. Table~\ref{tab:best_final} reports both for every system.

\begin{table}[h]
\centering
\caption{Composite scores of the final and best checkpoints of every system. The best checkpoint is identified after the run by scoring all checkpoints with the hidden graders. Six systems exceed the reference by final checkpoints and eight by best checkpoints.}
\label{tab:best_final}
\small
\begin{tabular}{@{}lcc@{}}
\toprule
System & Final checkpoint & Best checkpoint \\
\midrule
Claude Fable~5.1 & 0.515 & 0.531 \\
Kimi~K3 & 0.479 & 0.500 \\
GPT-5.6-Sol & 0.443 & 0.466 \\
GPT-6 Astra & 0.442 & 0.475 \\
Claude Opus~5 & 0.440 & 0.508 \\
GPT-5.5 & 0.403 & 0.404 \\
Gemini~3.8 Flash & 0.369 & 0.394 \\
DeepSeek V4.1 Flash & 0.351 & 0.452 \\
DeepSeek V4 Pro & 0.345 & 0.345 \\
MiniMax-M3 & 0.222 & 0.238 \\
GLM-5.2 & 0.188 & 0.228 \\
Qwen3-Max & 0.054 & 0.120 \\
\midrule
Reference model & 0.378 & 0.378 \\
\bottomrule
\end{tabular}
\end{table}

\subsection{Lifecycle and Isolation}

\paragraph{Lifecycle constants.}
The budget is $21{,}600$\,s of eligible time, counted from the start of the run and shared by the three phases. Eligible time is wall time minus credited pauses, and the clock stops while a credited pause lasts. During model development, the harness credits a pause when the provider reports an exhausted quota with a stated reset time, for the stated delay and for at most $12$ such events, or reports a transient service failure, for at most $600$\,s per event. A rate-limit error without a stated reset is not credited and consumes eligible time. Credited pauses are capped at six hours in total, and the wall time of a run is capped at $18$ hours. Phase A allows at most four attempts at the data check, each limited to $30$ minutes, and a run that fails all four ends there. Phase B ends when the agent ends its turn, and at the latest after $90$ minutes, followed by at most two repair rounds for scripts that fail to run. The diagnostics are then made read-only. In Phase C, the harness resumes the agent session whenever a turn ends, using the stall prompts of Appendix~\ref{app:task} after $20$, $40$, and $60$ minutes without a new checkpoint. A run ends early if an agent turn exits within $5$\,s, if three consecutive turns fail, or if eight consecutive turns end in an error despite changing the submission.

\paragraph{Isolation.}
Each run has its own container, which runs the agent as an unprivileged user and mounts only the working directory of the run. The graders, the reference model, and the observations after 2014 are never mounted, and grading happens after the run on the host. Agents work offline. The web search and browsing tools of every scaffold are disabled: web fetch and web search in Claude Code, web search in Codex, and the search, browsing, and subagent tools of Antigravity. The only permitted network use is installing Python packages, and audits of the transcripts found no other network use.

\subsection{Scaffold Identities and Reproducibility}

\paragraph{Scaffold identities.}
Table~\ref{tab:scaffolds} lists the scaffold and route of each system. The three additional runs of Claude Fable~5.1 in Section~\ref{sec:analysis_conditions} used the same configuration as its main run, with web access enabled through a filtering proxy in the two web conditions. Context management follows each scaffold's own behavior, and the largest contexts reached $658{,}258$ tokens for Claude Opus~5 and $314{,}562$ tokens for Kimi~K3.

\begin{table}[h]
\centering
\caption{Scaffolds and routes of the twelve systems. Systems routed through OpenRouter use one pinned provider each, given in parentheses, with no fallback.}
\label{tab:scaffolds}
\small
\begin{tabular}{@{}lll@{}}
\toprule
System & Scaffold & Route \\
\midrule
Claude Fable~5.1    & Claude Code 2.1.270 & Subscription \\
Kimi~K3             & Claude Code 2.1.223 & OpenRouter (Moonshot AI) \\
GPT-5.6-Sol         & Codex 0.146.1       & Subscription \\
GPT-6 Astra         & Codex 0.154.0       & Subscription \\
Claude Opus~5       & Claude Code 2.1.223 & Subscription \\
GPT-5.5             & Codex 0.146.1       & Subscription \\
Gemini~3.8 Flash    & Antigravity 1.2.0   & Subscription \\
DeepSeek V4.1 Flash & Claude Code 2.1.223 & OpenRouter (GMICloud) \\
DeepSeek V4 Pro     & Claude Code 2.1.223 & OpenRouter (GMICloud) \\
MiniMax-M3          & Claude Code 2.1.223 & OpenRouter (MiniMax) \\
GLM-5.2             & Claude Code 2.1.223 & OpenRouter (Z.ai) \\
Qwen3-Max           & Claude Code 2.1.223 & OpenRouter (Alibaba) \\
\bottomrule
\end{tabular}
\end{table}

\paragraph{Reproducibility.}
Every run used the same task files, prompts, graders, and container image, which runs Python~3.12 with NumPy~2.5.1 and SciPy~1.18.0. Each run records the configuration it ran under. The released code and data contain the harness, the graders, the final model of every system, the scores of every checkpoint, and the checksums that identify each of these components.
\section{Evaluation Details}
\label{app:evaluation}

This appendix describes how the graders evaluate a submitted model along the three dimensions introduced in Section~\ref{sec:evaluation}. For each dimension, we explain which aspect of ENSO it tests and why, how the graders run the submitted model to generate the results, and how these results are turned into a score between $0$ and $1$. The statistical and reconstruction tests follow the criteria used to validate the reference model~\citep{zhang2024physics}, and the forecast test is added for this benchmark. The three experiments use the model in three different roles: as a long free-running simulator, as the prior of a data-assimilation system, and as a forecast model. In all three, the graders call only the submitted time-stepping function, so the same equations and parameters are tested throughout. A model whose simulation does not remain finite is not scored. Because the task does not specify which calendar month corresponds to the start of model time, each model is evaluated under the calendar convention stated by its agent. The published reference model, which combines linear couplings with nonlinear drift and additive noise, is scored in the same way. It belongs to a richer model class than the linear stochastic models often used as ENSO baselines. The composite score is
\begin{equation}
C = 0.3\,S + 0.3\,D + 0.4\,P,
\end{equation}
where $S$, $D$, and $P$ are the scores for statistical properties, dynamical consistency, and predictive skill defined below. The number of systems that exceed the reference does not depend strongly on these weights. With equal weights the same six systems exceed it, with the weight on statistics raised to $0.4$ five do, and Claude Fable~5.1 and Kimi~K3 remain ahead until statistics carries more than about $94\%$ and $65\%$ of the weight, respectively, with the rest split equally between the other two dimensions.

\subsection{Statistical Properties}

ENSO is irregular, and its events differ in timing, strength, duration, and location. A stochastic model cannot, and is not expected to, reproduce the particular sequence of events in the observed record. What it should reproduce is the statistical behavior of ENSO, so we compare statistics of long simulations with those of the observations, choosing features that climate scientists use to characterize ENSO. The probability density function (PDF) of each temperature index measures the amplitude of variability and its asymmetry. Observed anomalies are positively skewed in the eastern Pacific, where the strongest El Ni\~no events occur, and negatively skewed in the central Pacific. The seasonal variance, computed separately for each calendar month, measures how ENSO is locked to the annual cycle. Events tend to peak near the end of the calendar year, so the observed variance is largest in December and smallest in spring or early summer. The autocorrelation function (ACF) measures how long anomalies persist and how quickly they reverse. The observed ACF becomes negative at lags beyond about one year, reflecting the alternation between El Ni\~no and La Ni\~na. Event frequencies measure ENSO diversity. Eastern Pacific (EP) and central Pacific (CP) El Ni\~no, extreme El Ni\~no, La Ni\~na, and multi-year events have different climate impacts, and a model with realistic variability can still produce the wrong mix of them. Finally, a bivariate diagnostic of event strength and location tests a nonlinear signature of diversity, namely that strong El Ni\~no events peak farther east than moderate ones.

\paragraph{Experiment.}
The graders run the submitted model freely for 2300 years from the zero state, with no observational input, so that its variability arises entirely from its own dynamics and stochastic forcing. The first 200 years are discarded so that the results do not depend on the initial state. The monthly values of $T_C$ and $T_E$ from the remaining 2100 years are divided into 30 non-overlapping 70-year segments, each as long as the observed record. The observed record consists of the Ni\~no~4 and Ni\~no~3 sea surface temperature anomaly indices from 1950 to 2019 used by the reference study~\citep{zhang2024physics}, taken from the data record of \citet{chen2022zenodo} and derived from HadISST1.1~\citep{rayner2003hadisst}. The PDF, seasonal variance, and ACF are computed for every segment and averaged over the 30 segments, which removes most of the sampling noise of a single 70-year record while keeping each statistic comparable with the observed one. In Figure~\ref{fig3_final_diagnostics}, the simulated curves show this segment average, and the shading around them shows one standard deviation across the 30 segments. Events are counted in 71-year segments and compared with the observed counts per 71 years reported by the reference study, using the same definitions. Each winter is classified from its December to February means of $T_E$ and $T_C$. It is an El Ni\~no if either mean exceeds $0.5\,^{\circ}$C, of EP type if $T_E$ is the larger and of CP type otherwise, and a La Ni\~na if either mean falls below $-0.5\,^{\circ}$C. An extreme El Ni\~no requires $T_E$ to reach $2.5\,^{\circ}$C, and multi-year events are consecutive El Ni\~no or La Ni\~na winters. For the bivariate diagnostic, each simulated El Ni\~no is mapped to an equatorial sea surface temperature profile with the regression of the reference study, which gives its strength as the peak anomaly and its location as the longitude of the peak. The observed events are located in the same way from the observed sea surface temperature field.

\paragraph{Score.}
For the PDF, the seasonal variance, and the ACF, the segment-averaged model curve $\mu$ is compared with the observed curve $o$ across its points $p$, which are anomaly values, calendar months, or lags,
\begin{equation}
s = \max\{0, r(\mu, o)\}\,\frac{\sum_p w_p \exp\!\left[-\frac{1}{2}\left(\frac{\mu_p - o_p}{0.35\,R}\right)^2\right]}{\sum_p w_p}.
\end{equation}
The first factor is the correlation between the two curves and measures whether the shape is right, for example whether the variance peaks in the correct month. The second factor measures whether the magnitudes agree, with a tolerance proportional to the variation $R$ of the observed curve across its points. PDF points are weighted by their probability, $w_p = \frac{1}{2}\left[\max(o_p, 0) + \max(\mu_p, 0)\right]$, and $w_p = 1$ for the other two curves. Each of these three groups averages the scores of $T_C$ and $T_E$. For each of the seven event types $k$, the mean simulated count $m_k$ is compared with the observed count $o_k$,
\begin{equation}
G_{\mathrm{event}} = \frac{1}{7}\sum_{k=1}^{7} \exp\!\left[-\frac{1}{2}\left(\frac{m_k - o_k}{0.2\,o_k}\right)^2\right],
\end{equation}
so a relative difference of $20\%$ lowers the contribution of that type to about $0.61$. The diversity group averages two criteria. The first requires simulated events that peak east of $150^{\circ}$W to be stronger on average than those that peak farther west,
\begin{equation}
G_{\mathrm{order}} = \min\left\{1, \max\left[0, \frac{\bar{S}_{\mathrm{EP}} - \bar{S}_{\mathrm{CP}}}{0.5\,^{\circ}\mathrm{C}}\right]\right\},
\end{equation}
where $\bar{S}_{\mathrm{EP}}$ and $\bar{S}_{\mathrm{CP}}$ are the mean strengths of the two sets. The second is the fraction of observed events that fall within the region of the strength and location plane occupied by $95\%$ of the simulated events. The statistical score combines the five groups $G_1, \dots, G_5$ as
\begin{equation}
S = \frac{1}{2}\cdot\frac{1}{5}\sum_{i=1}^{5} G_i + \frac{1}{2}\min_{i} G_i.
\end{equation}
The minimum term prevents a model that fails one feature of ENSO from compensating with the others.

\subsection{Dynamical Consistency}

Reproducing observed statistics does not establish that a model captures how its variables interact, because different dynamics can produce similar statistics. ENSO arises from coupling between the ocean and the atmosphere. Eastern Pacific warming weakens the trade winds, and the weaker winds drive eastward current anomalies and flatten the thermocline, and these changes in turn reinforce the warming. The slow recharge and discharge of equatorial heat content, represented by $h_W$, sets the timing of transitions between El Ni\~no and La Ni\~na. If a model represents these couplings correctly, observing some variables should reveal much about the others. We test this with data assimilation, the standard method for combining a model with incomplete observations. In Bayesian terms, the model provides a prior distribution of states, measurements of some variables update it to a posterior distribution, and the posterior mean of the unmeasured variables is their reconstruction. Because the unmeasured variables are informed only through the relationships encoded in the model, an accurate reconstruction is evidence that these relationships are realistic.

\paragraph{Experiment.}
The graders use the observed monthly indices from 1980 to 2014, the period available to the agents. An ensemble of 500 model states is advanced month by month by the submitted model, each member with its own stochastic forcing, so that the spread of the ensemble and the correlations among its variables reflect the model's own uncertainty and coupling. At the end of each month, an ensemble Kalman smoother assimilates the observed values of the conditioning variables, adjusting every member toward the observations in proportion to the covariances within the ensemble. The smoother also revises the states of the preceding three months with each new observation, so the reconstruction depends on the lagged relationships in the model, such as the lead of the western Pacific thermocline over eastern Pacific temperature, and not only on simultaneous ones. The ensemble mean of each unobserved variable is its reconstruction. The actual records of these variables are withheld from the smoother and used only for scoring. In test A, $T_C$ and $T_E$ are assimilated to reconstruct $h_W$, $u$, and $\tau$. In test B, the roles are reversed, so the coupling is tested in both directions.

\paragraph{Score.}
After the first two years, which allow the ensemble to adjust, each reconstruction $\hat{x}_t$ is compared with the withheld record $x_t$ by the Nash-Sutcliffe efficiency,
\begin{equation}
\mathrm{NSE} = \max\left\{0,\; 1 - \frac{\sum_t (\hat{x}_t - x_t)^2}{\sum_t (x_t - \bar{x})^2}\right\},
\end{equation}
where $\bar{x}$ is the mean of the record. The NSE is the fraction of the variance of the record that the reconstruction explains. Unlike a correlation, it also penalizes a reconstruction with the right timing but the wrong amplitude. The dynamical score $D$ is the mean of the five NSE values, three from test A and two from test B.

\subsection{Predictive Skill}

Forecasting is a primary practical use of ENSO models, and it is the only test on data that the agents never saw. The held-out period from 2015 to 2024 includes the extreme El Ni\~no of 2015/16, the multi-year La Ni\~na from 2020 to 2023, and the El Ni\~no of 2023/24, so it covers strong events, a multi-year event, and rapid transitions. Forecasts are compared with persistence, which assumes that the current anomaly continues unchanged. Persistence predicts no change, so a model that cannot improve on it has no forecast skill, and improving on it requires the model to capture how anomalies grow, decay, and reverse.

\paragraph{Experiment.}
From each month between January 2015 and December 2023, the graders initialize all five model variables with the observed values of that month and run a 50-member ensemble forward for 12 months without further observations. The members differ only in their stochastic forcing, and the ensemble mean serves as the forecast. The forecasts of $T_C$ and $T_E$ at lead times of 1 to 12 months, where the lead time is the interval between the start of a forecast and the month being predicted, are compared with the observed values for the target months. The persistence forecast for every lead time is the observed value of the starting month.

\paragraph{Score.}
For each index, let $\mathrm{RMSE}^{\mathrm{model}}_{\ell}$ and $\mathrm{RMSE}^{\mathrm{pers}}_{\ell}$ denote the root-mean-square errors of the model and of persistence over all forecasts at lead time $\ell$. The skill over all targets is
\begin{equation}
P^{\mathrm{all}} = \max\left\{0,\; 1 - \frac{\sum_{\ell=1}^{12} \mathrm{RMSE}^{\mathrm{model}}_{\ell}}{\sum_{\ell=1}^{12} \mathrm{RMSE}^{\mathrm{pers}}_{\ell}}\right\}.
\end{equation}
The extreme skill $P^{\mathrm{ext}}$ focuses on strong El Ni\~no and La Ni\~na conditions. It pools the forecasts at all start months and leads whose observed target anomaly exceeds one standard deviation in magnitude, and compares the root-mean-square errors of the model and of persistence over this pool,
\begin{equation}
P^{\mathrm{ext}} = \max\left\{0,\; 1 - \frac{\mathrm{RMSE}^{\mathrm{model}}_{\mathrm{ext}}}{\mathrm{RMSE}^{\mathrm{pers}}_{\mathrm{ext}}}\right\}.
\end{equation}
It is set to zero when fewer than ten such forecasts exist. The predictive score is
\begin{equation}
P = 0.7\,\overline{P^{\mathrm{all}}} + 0.3\,\overline{P^{\mathrm{ext}}},
\end{equation}
where the bars denote averages over $T_C$ and $T_E$. A score of zero means no improvement over persistence. The forecast correlations shown in Figure~\ref{fig3_final_diagnostics} are diagnostics and do not enter the score.
\section{Cost}
\label{app:infra}

For the systems routed through OpenRouter, the cost of a run is the sum of the billed cost of each model call, retrieved from OpenRouter by the identifier of the call. The systems on subscription plans pay a fixed fee, so their token use is reported instead, counted from their transcripts (Table~\ref{tab:cost}). Over all batches of runs, the OpenRouter spending was about $\$521$.

\begin{table}[h]
\centering
\caption{Model calls, prompt tokens, share of prompt tokens served from the provider's cache, and cost of the main run of each system. Subscription systems pay a fixed fee. The usage of GPT-5.6-Sol was not recorded. For DeepSeek V4.1 Flash, tokens are counted from the transcript and the cost is the change in account usage over its batch, in which it was the only system routed through OpenRouter.}
\label{tab:cost}
\small
\begin{tabular}{@{}llrrrr@{}}
\toprule
System & Route & Calls & Prompt tokens & Cached & Cost (USD) \\
\midrule
Kimi~K3             & OpenRouter & 954   & 120.3\,M & 99.2\% & 43.69 \\
GLM-5.2             & OpenRouter & 669   & 65.4\,M  & 98.0\% & 21.48 \\
Qwen3-Max           & OpenRouter & 3,250 & 308.1\,M & 95.3\% & 132.37 \\
DeepSeek V4 Pro     & OpenRouter & 589   & 48.7\,M  & 93.2\% & 7.67 \\
MiniMax-M3          & OpenRouter & 1,011 & 83.1\,M  & 98.2\% & 5.97 \\
DeepSeek V4.1 Flash & OpenRouter & 752   & 62.0\,M  & 98.4\% & 2.43 \\
\midrule
Claude Fable~5.1    & Subscription & 636   & 53.7\,M  & 97.2\% & fixed fee \\
Claude Opus~5       & Subscription & 446   & 133.9\,M & 99.4\% & fixed fee \\
GPT-5.5             & Subscription & 2,619 & 297.8\,M & 97.8\% & fixed fee \\
GPT-6 Astra         & Subscription & 432   & 42.7\,M  & 94.7\% & fixed fee \\
Gemini~3.8 Flash    & Subscription & 3,429 & 487.3\,M & 96.6\% & fixed fee \\
GPT-5.6-Sol         & Subscription & n/a   & n/a      & n/a    & fixed fee \\
\bottomrule
\end{tabular}
\end{table}
\clearpage
\section{Agent-Generated Models and Their Simplified Forms}
\label{app:models}
\begingroup
\allowdisplaybreaks[1]
\newcommand{\termcolor}[2]{#2}

This appendix presents Claude Fable~5.1 as a worked example of an original final model and its simplified form. The simplification procedure is described in Appendix~\ref{app:simplification}, and Table~\ref{tab:simplification_full} there summarizes the structure and scores of every model before and after simplification. The executable implementations, parameters, and model cards of all original and simplified models are provided in the released code and data.

\Needspace{12\baselineskip}

\subsection{Claude Fable 5.1}

\label{app:model-fable51}

\paragraph{Original model.} The model describes the coupled evolution of five state variables through seasonally modulated linear interactions and nonlinear feedbacks. Cubic damping constrains the amplitude of the simulated variability, while state-dependent stochastic forcing represents unresolved processes.

\begin{equation}
\label{eq:fable51-original}
\begin{aligned}
\frac{du}{dt} &= \termcolor{linear}{a_{11}u} + \termcolor{seasonal}{a_{12}(t)h_W} + \termcolor{linear}{a_{13}T_C} + \termcolor{linear}{a_{14}T_E} + \termcolor{linear}{a_{15}\tau} + \termcolor{noise}{\sigma_u(t)\dot W_u} \\
\frac{dh_W}{dt} &= \termcolor{linear}{a_{21}u} + \termcolor{seasonal}{a_{22}(t)h_W} + \termcolor{linear}{a_{23}T_C} + \termcolor{linear}{a_{24}T_E} + \termcolor{linear}{a_{25}\tau} + \termcolor{nonlinear}{a_{26}\frac{T_E^2}{1+(T_E/T_{s,h})^2}} + \termcolor{nonlinear}{a_{27}h_W^3} \\
&\quad + \termcolor{offset}{a_{28}} + \termcolor{noise}{\sigma_h(h_W,t)\dot W_h} \\
\frac{dT_C}{dt} &= \termcolor{linear}{a_{31}u} + \termcolor{seasonal}{a_{32}(t)h_W} + \termcolor{seasonal}{a_{33}(t)T_C} + \termcolor{linear}{a_{34}T_E} + \termcolor{linear}{a_{35}\tau} + \termcolor{nonlinear}{a_{36}T_C^2} + \termcolor{nonlinear}{a_{37}T_C^2\,\mathbf{1}_{T_C>0}} \\
&\quad + \termcolor{nonlinear}{a_{38}T_C^3} + \termcolor{offset}{a_{39}} + \termcolor{noise}{\sigma_C(T_C)\dot W_C} \\
\frac{dT_E}{dt} &= \termcolor{linear}{a_{41}u} + \termcolor{seasonal}{a_{42}(t)h_W} + \termcolor{linear}{a_{43}T_C} + \termcolor{seasonal}{a_{44}(t)T_E} + \termcolor{linear}{a_{45}\tau} + \termcolor{nonlinear}{a_{46}(t)\frac{T_E^2}{1+(T_E/T_{s,E})^2}} \\
&\quad + \termcolor{nonlinear}{a_{47}T_E^3} + \termcolor{offset}{a_{48}} + \termcolor{noise}{\sigma_E(T_E,t)\dot W_E} \\
\frac{d\tau}{dt} &= \termcolor{linear}{a_{51}u} + \termcolor{linear}{a_{52}h_W} + \termcolor{linear}{a_{53}T_C} + \termcolor{linear}{a_{54}T_E} + \termcolor{linear}{a_{55}\tau} + \termcolor{noise}{\sigma_{\tau}(T_C,t)\dot W_{\tau}}
\end{aligned}
\end{equation}

Here, $a_{ij}$ denotes the coefficient of the corresponding term. Coefficients that depend on $t$ vary seasonally. The parameters $T_{s,h}$ and $T_{s,E}$ set the saturation scales of the nonlinear feedbacks. The indicator $\mathbf{1}_{\{T_C>0\}}$ activates the associated term only for positive $T_C$. The terms $\dot W_u$, $\dot W_h$, $\dot W_C$, $\dot W_E$, and $\dot W_\tau$ represent independent Gaussian white noise. The corresponding $\sigma$ functions determine the noise amplitudes and their dependence on season or model state.

\paragraph{Simplified model.}
The simplified model removes all nonlinear drift terms and confines seasonal variation in the drift to the growth and damping of $T_C$ and $T_E$. It retains a refitted linear coupling among the five variables and preserves the original state-dependent stochastic forcing. The resulting model contains 34 drift coefficients. Relative to the original model, it improves the statistical properties and predictive skill but lowers dynamical consistency, revealing a trade-off among the evaluation criteria.

\begin{equation}
\label{eq:fable51-simplified}
\begin{aligned}
\frac{du}{dt} &= {\termcolor{linear}{\widetilde a_{11}u}} + {\termcolor{linear}{\widetilde a_{12}h_W}} + {\termcolor{linear}{\widetilde a_{13}T_C}} + {\termcolor{linear}{\widetilde a_{14}T_E}} + {\termcolor{linear}{\widetilde a_{15}\tau}} + {\termcolor{noise}{\sigma_u(t)\dot W_u}} \\
\frac{dh_W}{dt} &= {\termcolor{linear}{\widetilde a_{21}u}} + {\termcolor{linear}{\widetilde a_{22}h_W}} + {\termcolor{linear}{\widetilde a_{23}T_C}} + {\termcolor{linear}{\widetilde a_{24}T_E}} + {\termcolor{linear}{\widetilde a_{25}\tau}} + {\termcolor{offset}{\widetilde a_{28}}} + {\termcolor{noise}{\sigma_h(h_W,t)\dot W_h}} \\
\frac{dT_C}{dt} &= {\termcolor{linear}{\widetilde a_{31}u}} + {\termcolor{linear}{\widetilde a_{32}h_W}} + {\termcolor{seasonal}{\widetilde a_{33}(t)T_C}} + {\termcolor{linear}{\widetilde a_{34}T_E}} + {\termcolor{linear}{\widetilde a_{35}\tau}} + {\termcolor{offset}{\widetilde a_{39}}} + {\termcolor{noise}{\sigma_C(T_C)\dot W_C}} \\
\frac{dT_E}{dt} &= {\termcolor{linear}{\widetilde a_{41}u}} + {\termcolor{linear}{\widetilde a_{42}h_W}} + {\termcolor{linear}{\widetilde a_{43}T_C}} + {\termcolor{seasonal}{\widetilde a_{44}(t)T_E}} + {\termcolor{linear}{\widetilde a_{45}\tau}} + {\termcolor{offset}{\widetilde a_{48}}} + {\termcolor{noise}{\sigma_E(T_E,t)\dot W_E}} \\
\frac{d\tau}{dt} &= {\termcolor{linear}{\widetilde a_{51}u}} + {\termcolor{linear}{\widetilde a_{52}h_W}} + {\termcolor{linear}{\widetilde a_{53}T_C}} + {\termcolor{linear}{\widetilde a_{54}T_E}} + {\termcolor{linear}{\widetilde a_{55}\tau}} + {\termcolor{noise}{\sigma_{\tau}(T_C,t)\dot W_{\tau}}}
\end{aligned}
\end{equation}

Here, $\widetilde a_{ij}$ denotes a coefficient refitted for the simplified model. The subscripts are retained from the original formulation to show the correspondence between terms in the two models. The coefficients $\widetilde a_{33}(t)$ and $\widetilde a_{44}(t)$ vary seasonally, while the remaining $\widetilde a_{ij}$ are constant. The noise processes and the functions $\sigma_u$, $\sigma_h$, $\sigma_C$, $\sigma_E$, and $\sigma_\tau$ are unchanged from the original model.

\Needspace{7\baselineskip}

\begin{center}

\begin{tabular}{lrrrr}

\toprule Model & Stat. & Dyn. & Pred. & Composite \\ \midrule

Original & 0.715 & 0.639 & 0.274 & 0.515 \\

Simplified & 0.810 & 0.600 & 0.297 & 0.542 \\

\bottomrule

\end{tabular}

\end{center}

\Needspace{9\baselineskip}
\paragraph{Original parameters and auxiliary definitions.}
\begin{longtable}{@{}p{0.15\linewidth}p{0.30\linewidth}p{0.15\linewidth}p{0.30\linewidth}@{}}
\toprule
Symbol & Value & Symbol & Value \\ \midrule
\endhead
$a_{11}$ & $-0.9170$ & $a_{13}$ & $-0.4020$ \\
$a_{14}$ & $0.2600$ & $a_{15}$ & $0.1140$ \\
$a_{21}$ & $-0.1640$ & $a_{23}$ & $0.09686$ \\
$a_{24}$ & $-0.1500$ & $a_{25}$ & $-0.07482$ \\
$a_{26}$ & $-0.3000$ & $a_{27}$ & $-1.184$ \\
$a_{28}$ & $0.004840$ & $T_{s,h}$ & $0.4000$ \\
$a_{31}$ & $0.1628$ & $a_{34}$ & $0.1210$ \\
$a_{35}$ & $0.08530$ & $a_{36}$ & $-1.020$ \\
$a_{37}$ & $-0.7500$ & $a_{38}$ & $-7.181$ \\
$a_{39}$ & $0.004000$ & $a_{41}$ & $0.2403$ \\
$a_{42}(t)$ & $0.05899 +\allowbreak 0.1000 s_1$ & $a_{43}$ & $-0.1399$ \\
$a_{45}$ & $0.1285$ & $a_{46}(t)$ & $0.5050 - 0.6060 s_1$ \\
$a_{47}$ & $-1.038$ & $a_{48}$ & $-0.006260$ \\
$T_{s,E}$ & $0.3500$ & $a_{51}$ & $0.3296$ \\
$a_{52}$ & $-0.2758$ & $a_{53}$ & $3.062$ \\
$a_{54}$ & $-1.118$ & $a_{55}$ & $-0.9918$ \\
$S_T$ & $7.500\ \text{degC}$ & $S_h$ & $150.0\ \text{m}$ \\
$S_u$ & $1.500\ \text{m/s}$ & $S_{\tau}$ & $5.000\ \text{m/s}$ \\
$S_t$ & $2.000\ \text{months}$ & $x_{\max}$ & $3.000$ \\
\bottomrule
\end{longtable}

The bound $x_{\max}=3$ clips the nondimensional state at $\pm 3$, corresponding to $\pm 22.5\,^{\circ}$C in temperature and $\pm 450$\,m in thermocline depth, far outside the observed range, so it acts only as a numerical safeguard and does not alter the linear drift.

\begin{longtable}{@{}p{0.22\linewidth}p{0.74\linewidth}@{}}
\toprule
Symbol & Definition or value \\ \midrule
\endhead
$a_{12}(t)$ & $0.8500 +\allowbreak 0.1700 c_1 +\allowbreak 0.4250 s_1$ \\
$a_{22}(t)$ & $-0.1640 +\allowbreak 0.1044 c_1 - 0.08738 s_1$ \\
$a_{32}(t)$ & $0.1073 +\allowbreak 0.01866 c_1 +\allowbreak 0.03000 s_1$ \\
$a_{33}(t)$ & $-0.2500 +\allowbreak 0.09000 c_1 - 0.2400 s_1$ \\
$a_{44}(t)$ & $-0.1510 - 0.006228 c_1 - 0.2700 s_1 - 0.1200 c_2 - 0.1500 s_2$ \\
$\sigma_u(t)$ & $0.09400\,\allowbreak (1 - 0.1000 c_1 +\allowbreak 0.03000 s_1 - 0.1500 c_2 +\allowbreak 0.1500 s_2)$ \\
$\sigma_h(h_W,t)$ & $0.02400\,\allowbreak (1 +\allowbreak 0.2000 c_1)\max(1 - 4.000\,\allowbreak h_W,\allowbreak \,\allowbreak 0.2)$ \\
$\sigma_C(T_C)$ & $0.03000\max(1 - 2.000\,\allowbreak T_C,\allowbreak \,\allowbreak 0.2)\max(1 - 30.00\,\allowbreak T_C^2,\allowbreak \,\allowbreak 0.2)$ \\
$\sigma_E(T_E,t)$ & $0.03700\,\allowbreak (1 - 0.6000 c_1 +\allowbreak 0.4000 s_1 - 0.3000 c_2)\min(\max(1 +\allowbreak 5.500\,\allowbreak T_E,\allowbreak \,\allowbreak 0.2),\allowbreak \,\allowbreak 2.500)$ \\
$\sigma_{\tau}(T_C,t)$ & $0.4500\,\allowbreak (1 +\allowbreak 0.1300 c_1 - 0.05000 s_1 - 0.08000 c_2 +\allowbreak 0.2000 s_2 - 0.2000 c_3 - 0.1000 s_3)\max(1 +\allowbreak 0.8000\,\allowbreak T_C,\allowbreak \,\allowbreak 0.2)$ \\
$t_m$ & $2t \ (\text{months since 1 Jan of the run start})$ \\
$c_1$ & $\cos\!\big(2\pi\cdot 1\,\allowbreak (t_m-0.5)/12\big)$ \\
$s_1$ & $\sin\!\big(2\pi\cdot 1\,\allowbreak (t_m-0.5)/12\big)$ \\
$c_2$ & $\cos\!\big(2\pi\cdot 2\,\allowbreak (t_m-0.5)/12\big)$ \\
$s_2$ & $\sin\!\big(2\pi\cdot 2\,\allowbreak (t_m-0.5)/12\big)$ \\
$c_3$ & $\cos\!\big(2\pi\cdot 3\,\allowbreak (t_m-0.5)/12\big)$ \\
$s_3$ & $\sin\!\big(2\pi\cdot 3\,\allowbreak (t_m-0.5)/12\big)$ \\
\bottomrule
\end{longtable}

\Needspace{9\baselineskip}
\paragraph{Reduced drift coefficients.}
\begin{longtable}{@{}p{0.15\linewidth}p{0.30\linewidth}p{0.15\linewidth}p{0.30\linewidth}@{}}
\toprule
Symbol & Value & Symbol & Value \\ \midrule
\endhead
$\widetilde a_{11}$ & $-0.8523968$ & $\widetilde a_{12}$ & $0.9358338$ \\
$\widetilde a_{13}$ & $-0.4152634$ & $\widetilde a_{14}$ & $0.2980273$ \\
$\widetilde a_{15}$ & $0.1134559$ & $\widetilde a_{21}$ & $-0.1834501$ \\
$\widetilde a_{22}$ & $-0.2025177$ & $\widetilde a_{23}$ & $0.1094091$ \\
$\widetilde a_{24}$ & $-0.2023363$ & $\widetilde a_{25}$ & $-0.07096293$ \\
$\widetilde a_{31}$ & $0.1524918$ & $\widetilde a_{32}$ & $0.1193508$ \\
$\widetilde a_{34}$ & $0.09620686$ & $\widetilde a_{35}$ & $0.0811469$ \\
$\widetilde a_{41}$ & $0.2445608$ & $\widetilde a_{42}$ & $0.07082664$ \\
$\widetilde a_{43}$ & $-0.1592483$ & $\widetilde a_{45}$ & $0.1280047$ \\
$\widetilde a_{51}$ & $0.3361003$ & $\widetilde a_{52}$ & $-0.2772871$ \\
$\widetilde a_{53}$ & $3.029652$ & $\widetilde a_{54}$ & $-1.102882$ \\
$\widetilde a_{55}$ & $-0.9893288$ & $\widetilde a_{28}$ & $0.00227347$ \\
$\widetilde a_{39}$ & $-0.003035909$ & $\widetilde a_{48}$ & $-0.00205137$ \\
\bottomrule
\end{longtable}
\begin{longtable}{@{}p{0.22\linewidth}p{0.74\linewidth}@{}}
\toprule
Symbol & Definition or value \\ \midrule
\endhead
$\widetilde a_{33}$ & $-0.3096332 +\allowbreak 0.03899662 c_1 - 0.2438837 s_1$ \\
$\widetilde a_{44}$ & $-0.1736369 - 0.02538335 c_1 - 0.3242342 s_1 - 0.123943 c_2 - 0.1380168 s_2$ \\
\bottomrule
\end{longtable}

All noise definitions and numerical bounds in the original parameter table also apply to the reduced model.

\endgroup
\section{Claude Fable~5.1 under Four Conditions}
\label{app:fable}

Table~\ref{tab:fable_conditions} gives the scores of the four Claude Fable~5.1 runs discussed in Section~\ref{sec:analysis_conditions}.

\begin{table}[h]
\centering
\newcommand{\dup}[1]{\,{\scriptsize\textcolor{gaincolor}{($\uparrow$#1)}}}
\newcommand{\ddn}[1]{\,{\scriptsize\textcolor{dropcolor}{($\downarrow$#1)}}}
\caption{Claude Fable~5.1 under four conditions, each run once for six hours of eligible time. Best is the highest-scoring checkpoint and final is the last one, and the three component scores are for the best checkpoint. Parentheses give the change relative to the offline run (green for an increase, red for a decrease), and bold marks the best value in each column.}
\label{tab:fable_conditions}
\small
\setlength{\tabcolsep}{3pt}
\begin{tabular}{@{}llccccc@{}}
\toprule
& & \multicolumn{2}{c}{Composite} & \multicolumn{3}{c}{Best checkpoint} \\
\cmidrule(lr){3-4}\cmidrule(lr){5-7}
Condition & Access & Best & Final & Stat. & Dyn. & Pred. \\
\midrule
Offline (formal) & None             & \textbf{0.531} & \textbf{0.515} & 0.741 & \textbf{0.632} & 0.297 \\
Years anonymized & None             & 0.482\ddn{0.049} & 0.463\ddn{0.052} & 0.656\ddn{0.085} & 0.576\ddn{0.056} & 0.281\ddn{0.016} \\
Web access 1     & Literature, data & 0.475\ddn{0.056} & 0.386\ddn{0.129} & 0.638\ddn{0.103} & 0.572\ddn{0.060} & 0.280\ddn{0.017} \\
Web access 2     & Literature       & 0.528\ddn{0.003} & 0.406\ddn{0.109} & \textbf{0.798}\dup{0.057} & 0.556\ddn{0.076} & \textbf{0.306}\dup{0.009} \\
\bottomrule
\end{tabular}
\end{table}

\section{Structural Simplification of All Twelve Models}
\label{app:simplification}

Groups of terms were removed from each final model, the remaining coefficients were refitted to long simulations of the original model, and the reduced model was rescored with the same graders and random seed. A reduction was kept if it retained at least $95\%$ of the original composite score and $90\%$ of every dimension score, and if it kept the model's ENSO complexity, in particular the asymmetry of its temperature distribution, which the scores alone can miss. For GPT-6 Astra and the two DeepSeek models, removing both nonlinear drift and multiplicative noise keeps the scores within these limits but leaves a nearly symmetric distribution and lower event diversity, so removal stopped before the asymmetry was lost. Asymmetry was judged by comparing the simulated distribution of $T_E$ with the observed one, which is positively skewed (Figure~\ref{fig3_final_diagnostics}e), and a reduction whose distribution became nearly symmetric was rejected. When reductions in both families qualified, the higher-scoring one was kept. The three models that score below $0.24$ already miss several statistical features of ENSO, so their reductions were screened on the scores alone. Table~\ref{tab:simplification_full} extends Table~\ref{tab:simplification} to all twelve final models, including these three, which are not part of the mechanism analysis in Section~\ref{sec:analysis_mechanism}.

\begin{table}[h]
\centering
\newcommand{\dup}[1]{\,{\scriptsize\textcolor{gaincolor}{($\uparrow$#1)}}}
\newcommand{\ddn}[1]{\,{\scriptsize\textcolor{dropcolor}{($\downarrow$#1)}}}
\newcommand{\dzero}{\,{\scriptsize\textcolor{gray}{(0.000)}}}
\caption{Structure and scores of all twelve final models before and after simplification, in the format of Table~\ref{tab:simplification}. The three models below the middle rule score below $0.24$ and already miss several statistical features of ENSO, so their reductions were screened on the scores alone and are not included in the mechanism analysis.}
\label{tab:simplification_full}
\small
\setlength{\tabcolsep}{3.5pt}
\begin{adjustbox}{max width=\linewidth}
\begin{tabular}{@{}llccllll@{}}
\toprule
\multirow{2}{*}{Model} & \multirow{2}{*}{Version} & \multicolumn{2}{c}{Structure} & \multicolumn{3}{c}{Component scores} & \multirow{2}{*}{Composite} \\
\cmidrule(lr){3-4}\cmidrule(lr){5-7}
 & & Drift & Noise & Stat. & Dyn. & Pred. & \\
\midrule
\multirow{2}{*}{Claude Fable~5.1} & Original & NL & Mult. & 0.715 & \textbf{0.639} & 0.274 & 0.515 \\
 & Simplified & L & Mult. & \textbf{0.810}\dup{0.095} & 0.600\ddn{0.039} & \textbf{0.297}\dup{0.023} & \textbf{0.542}\dup{0.027} \\
\addlinespace[2pt]
\multirow{2}{*}{Kimi~K3} & Original & NL & Mult. & \textbf{0.620} & 0.570 & \textbf{0.305} & \textbf{0.479} \\
 & Simplified & L & Mult. & 0.575\ddn{0.045} & \textbf{0.573}\dup{0.003} & 0.302\ddn{0.003} & 0.465\ddn{0.014} \\
\addlinespace[2pt]
\multirow{2}{*}{GPT-5.6-Sol} & Original & NL & Add. & 0.554 & 0.515 & \textbf{0.306} & 0.443 \\
 & Simplified & NL & Add. & \textbf{0.613}\dup{0.059} & \textbf{0.529}\dup{0.014} & 0.304\ddn{0.002} & \textbf{0.464}\dup{0.021} \\
\addlinespace[2pt]
\multirow{2}{*}{GPT-6 Astra} & Original & NL & Add. & 0.530 & \textbf{0.629} & 0.237 & 0.442 \\
 & Simplified & NL & Add. & \textbf{0.598}\dup{0.068} & 0.611\ddn{0.018} & \textbf{0.248}\dup{0.011} & \textbf{0.462}\dup{0.020} \\
\addlinespace[2pt]
\multirow{2}{*}{Claude Opus~5} & Original & NL & Mult. & 0.548 & \textbf{0.589} & 0.247 & 0.440 \\
 & Simplified & NL & Add. & \textbf{0.629}\dup{0.081} & 0.567\ddn{0.022} & \textbf{0.286}\dup{0.039} & \textbf{0.473}\dup{0.033} \\
\addlinespace[2pt]
\multirow{3}{*}{GPT-5.5} & Original & NL & Mult. & \textbf{0.412} & 0.595 & 0.252 & \textbf{0.403} \\
 & Simplified & NL & Add. & 0.390\ddn{0.022} & \textbf{0.600}\dup{0.005} & 0.251\ddn{0.001} & 0.398\ddn{0.005} \\
 & Simplified & L & Mult. & 0.387\ddn{0.025} & 0.590\ddn{0.005} & \textbf{0.255}\dup{0.003} & 0.395\ddn{0.008} \\
\addlinespace[2pt]
\multirow{2}{*}{Gemini~3.8 Flash} & Original & NL & Mult. & \textbf{0.406} & \textbf{0.503} & 0.241 & \textbf{0.369} \\
 & Simplified & NL & Add. & 0.389\ddn{0.017} & 0.499\ddn{0.004} & 0.241\dzero & 0.362\ddn{0.007} \\
\addlinespace[2pt]
\multirow{2}{*}{\makecell[l]{DeepSeek\\V4.1 Flash}} & Original & NL & Mult. & 0.418 & 0.508 & 0.183 & 0.351 \\
 & Simplified & L & Mult. & \textbf{0.419}\dup{0.001} & \textbf{0.559}\dup{0.051} & \textbf{0.244}\dup{0.061} & \textbf{0.391}\dup{0.040} \\
\addlinespace[2pt]
\multirow{2}{*}{\makecell[l]{DeepSeek\\V4 Pro}} & Original & NL & Mult. & \textbf{0.285} & 0.530 & \textbf{0.251} & \textbf{0.345} \\
 & Simplified & L & Mult. & 0.269\ddn{0.016} & \textbf{0.532}\dup{0.002} & 0.250\ddn{0.001} & 0.340\ddn{0.005} \\
\midrule
\multirow{2}{*}{MiniMax-M3} & Original & NL & Mult. & \textbf{0.231} & \textbf{0.326} & 0.136 & \textbf{0.222} \\
 & Simplified & NL & Mult. & 0.229\ddn{0.002} & 0.323\ddn{0.003} & 0.136\dzero & 0.220\ddn{0.002} \\
\addlinespace[2pt]
\multirow{2}{*}{GLM-5.2} & Original & NL & Mult. & 0.222 & 0.261 & 0.108 & 0.188 \\
 & Simplified & L & Add. & \textbf{0.241}\dup{0.019} & \textbf{0.381}\dup{0.120} & \textbf{0.265}\dup{0.157} & \textbf{0.293}\dup{0.105} \\
\addlinespace[2pt]
\multirow{2}{*}{Qwen3-Max} & Original & NL & Add. & 0.118 & 0.062 & 0.000 & 0.054 \\
 & Simplified & L & Add. & 0.118\dzero & 0.062\dzero & 0.000\dzero & 0.054\dzero \\
\bottomrule
\end{tabular}
\end{adjustbox}
\end{table}
\section{Further Related Work}
\label{app:related}

\paragraph{Equation discovery.} Data-driven methods identify governing equations by sparse regression over candidate terms~\citep{brunton2016sindy} or by symbolic regression~\citep{schmidt2009distilling}, and language models now propose equation skeletons that are fitted and scored against data~\citep{shojaee2025llmsr,shojaee2025llmsrbench}. These methods work on prepared data within a fixed pipeline, whereas agents in \bench{} process raw observations, choose what to check, and are judged on behavior that a good fit does not guarantee. Machine-learning emulators of climate have their own benchmarks~\citep{watsonparris2022climatebench}.

\paragraph{Conceptual ENSO models.} Low-order ENSO models range from the recharge oscillator~\citep{jin1997} to stochastic multiscale models of ENSO diversity~\citep{chen2022multiscale}, an extended recharge oscillator that forecasts competitively~\citep{zhao2024xro}, and an automated framework that builds stochastic models of ENSO diversity by selecting their terms with causal inference~\citep{zhang2024physics}. That study validated its models by their statistics, their event diversity, and their recovery of unobserved variables through data assimilation. \bench{} adopts these criteria for its first two dimensions and adds forecasts of held-out years. Its contribution lies in the workflow around these tests, in which general-purpose agents process raw observations, write and freeze their own diagnostics, and develop a model over hours without seeing any score. Whether ENSO asymmetry arises from nonlinear dynamics or multiplicative noise remains debated~\citep{martinezvillalobos2019,srinivas2024,vialard2025}. The agent-built models of this study add candidates to the asymmetry debate under common tests.
\end{document}